\documentclass{article} 
\usepackage{iclr2027_conference,times}
\usepackage{hyperref}
\usepackage{url}
\usepackage{booktabs}
\usepackage{tabularx}
\usepackage[utf8]{inputenc} 
\usepackage[T1]{fontenc}    
\usepackage{hyperref}       
\usepackage{url}            
\usepackage{booktabs}       
\usepackage{amsfonts}       
\usepackage{nicefrac}       
\usepackage{microtype}      
\usepackage{xcolor}         
\usepackage{amsmath}
\usepackage{amssymb}
\usepackage{mathtools}
\usepackage{amsthm}
\usepackage{multirow}
\usepackage{float}
\usepackage{algorithm}
\usepackage{threeparttable}
\usepackage{algorithmic}
\usepackage{graphicx}
\usepackage{subfig}
\usepackage[export]{adjustbox}
\usepackage{caption}
\usepackage{wrapfig}
\usepackage{setspace}
\usepackage[misc]{ifsym}
\usepackage{amssymb}
\usepackage{amsmath}
\usepackage{listings}
\usepackage{xcolor}
\usepackage{colortbl}
\usepackage{pifont}
\usepackage{makecell}  
\usepackage[most]{tcolorbox}

\definecolor{tablelightblue}{HTML}{E1E8FF}
\definecolor{darkred}{RGB}{177,0,27}

\definecolor{tablelightblue}{HTML}{E1E8FF}
\definecolor{groupgray}{gray}{0.92}
\definecolor{visualblue}{RGB}{104,157,246}
\definecolor{mappingyellow}{RGB}{251,201,54}
\definecolor{integrationgreen}{RGB}{93,185,117}
\definecolor{disentanglered}{RGB}{238,105,93}
\definecolor{othergray}{RGB}{165,165,165}

\title{PhysFieldBench: Can Multimodal Models \\ Understand Physical Fields?}
\author{
{\bfseries Yuezhou Ma$^*$ \quad Huikun Weng$^*$ \quad Jialong Wu \quad Chenyi Zhao}\\
{\bfseries Hang Zhou \quad Haonan Shangguan \quad Jianmin Wang \quad Mingsheng Long$^\dagger$}\\[2pt]
School of Software, BNRist, Tsinghua University, China\\[2pt]
\texttt{\{mayz24,wenghk22\}@mails.tsinghua.edu.cn}\\
\texttt{\{jimwang,mingsheng\}@tsinghua.edu.cn}\\[2pt]
{\small $^*$Equal contribution. \quad $^\dagger$Corresponding author.}
}

\begin{document}
\maketitle
\begin{abstract}
Multimodal large language models (MLLMs) are increasingly envisioned as core components of scientific and engineering agents, yet their ability to interpret physical fields remains poorly understood. Existing physics benchmarks largely emphasize textbook problem solving or intuitive physical reasoning, leaving open whether MLLMs can infer physically meaningful information from continuous field observations. We introduce \emph{PhysFieldBench}, a benchmark comprising 24 tasks and 1,160 evaluation examples across controlled equation fields, simulated physical fields, and observed physical fields. The tasks assess three forms of inference: identifying physical mechanisms, comparing latent control variables, and predicting outcome properties. Across representative open-source and proprietary MLLMs, zero-shot performance is low: the best model achieves a chance-normalized score of 29.3, while several open-source models remain near chance. In contrast, a task-specific supervised vision transformer performs substantially better, demonstrating that the inputs contain learnable physical information. To diagnose these failures, a structured self-explanation analysis attributes most errors to missed visual patterns and incorrect visual-to-physical mappings. Further, to explore whether post-training can improve physical inference and generalize to unseen tasks, we compare supervised fine-tuning with final answers or chain-of-thought supervision and reinforcement learning. Final-answer supervision performs best overall but transfers less effectively, whereas reinforcement learning after chain-of-thought supervision achieves the best generalization. Together, these findings highlight the need to improve visual-to-physical grounding and cross-task generalization for MLLMs to reliably interpret physical fields in scientific and engineering workflows.
\end{abstract}
\section{Introduction}
Multimodal large language models (MLLMs) have achieved strong performance across general vision-language tasks
\citep{comanici2025gemini25,bai2025qwen3vl,liu2023visualinstruction,openai2023gpt4}.
Building on these advances, they are increasingly envisioned as core components of scientific and engineering agents
\citep{wang2023scientificdiscovery,boiko2023autonomous,bran2024augmenting}.
To operate reliably in such settings, these agentic models must go beyond recognizing explicit visual content to interpret the structures, dynamics, and latent physical factors encoded in visual observations.
Moreover, critical information in scientific and engineering workflows is often represented as physical fields produced by simulations, experiments, and real-world observations \citep{NEURIPS2025_332b4fbe,zhou2024unisolver,lam2023graphcast}, rather than discrete objects, natural scenes, or textbook-style diagrams.
Therefore, evaluating whether MLLMs can extract physically meaningful information from the spatiotemporal patterns of such fields is essential for assessing their capabilities in these agentic workflows.
\begin{figure*}[t]
\centering
\includegraphics[width=\columnwidth]{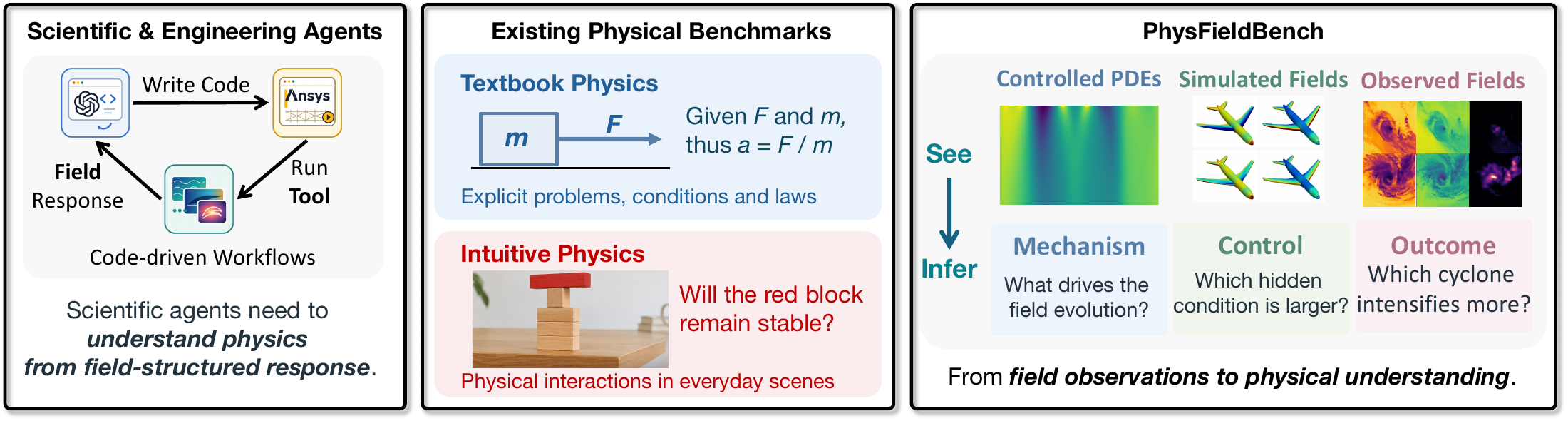}
\caption{While existing physics benchmarks focus primarily on textbook problems or intuitive reasoning, PhysFieldBench targets the physical-field understanding required by scientific and engineering agents: inferring mechanisms, control variables, and outcomes from diverse physical fields.}
\label{fig:Introduction}
\vspace{-15pt}
\end{figure*}

Existing physics-related benchmarks largely follow two paradigms. The first focuses on textbook-style scientific problem solving, presenting models with textual descriptions, equations, numerical variables, or diagrams and evaluating their ability to apply explicitly provided physical laws and conditions
\citep{xiang2026seephys}. The second studies intuitive physics in object-centric scenes, testing whether models can reason about motion, collision, stability, support relations, and interaction outcomes
\citep{chow2025physbench}. While these benchmarks have substantially advanced the evaluation of symbolic and intuitive physical reasoning, their inputs and targets differ from those encountered in many scientific and engineering workflows. In such workflows, models often receive continuous physical fields obtained from simulations, experiments, or real-world observations. To support downstream analysis, prediction, and decision-making, an agent must extract and use the physical information encoded in these fields rather than merely recognize their visual appearance. Existing multimodal benchmarks therefore provide limited evaluation of this physical-field understanding.

To formalize this capability, we define \emph{physical-field inference} as a multimodal task in which models infer physically grounded targets from one or more field observations. Unlike textbook-style problems, the input does not explicitly specify the physical mechanism, control variable, or outcome of interest; models must instead relate visual field patterns to attributes determined by the underlying physical process. In our benchmark, these targets are grounded in governing equations, simulation metadata, solver outputs, or observational annotations. We organize the tasks along three complementary axes, illustrated in Figure~\ref{fig:Introduction}. \emph{Mechanism inference} concerns identifying the physical processes that govern field evolution, such as diffusion, convection, reaction, or their coupled effects. \emph{Control inference} concerns comparing latent physical conditions or control variables, such as Reynolds number, Mach number, diffusion coefficient, cooling time, or vibration frequency. \emph{Outcome inference} concerns predicting system-level properties or future changes, such as drag, lift, tropical cyclone intensification, or precipitation coverage. Together, these axes provide an operational framework for evaluating whether MLLMs can interpret and use the physical information encoded in continuous fields.

To instantiate this framework, we introduce \emph{PhysFieldBench}, a benchmark for evaluating physical-field understanding in MLLMs. It comprises 24 tasks and 1,160 evaluation examples from nine data sources, spanning controlled equation fields, simulated physical fields, and observed physical fields. The benchmark covers diverse field representations, including field images, temporal montages, multi-view visualizations, and multimodal observation panels, and is organized around the three inference axes described above. Across a broad range of open-source and proprietary MLLMs, zero-shot performance remains low, while a task-specific supervised vision transformer performs substantially better, indicating that the rendered observations contain learnable task-relevant physical signals. These results motivate two follow-up analyses. To characterize the sources of zero-shot failures, we conduct a structured self-explanation analysis, which attributes recurring errors to missed visual patterns and incorrect visual-to-physical mappings. To examine whether domain-specific post-training can improve physical-field inference and transfer beyond training-covered tasks, we compare final-answer supervision, chain-of-thought supervision, and reinforcement learning. Final-answer supervision achieves the strongest overall performance but transfers less effectively to held-out tasks, whereas reinforcement learning following chain-of-thought supervision achieves the best held-out-task performance. Together, these findings reveal a substantial gap between general multimodal perception and reliable physical-field inference, motivating PhysFieldBench as a diagnostic benchmark for scientific and engineering agents. Overall, our contributions can be summarized as follows:
\begin{itemize}
    \item We formulate \emph{physical-field inference} as a multimodal evaluation setting for \emph{physical-field understanding} required by scientific and engineering agents to interpret field observations.
    \item We introduce \textbf{PhysFieldBench}, comprising 24 tasks and 1,160 evaluation examples from nine data sources across controlled equation fields, simulated physical fields, and observed physical fields, organized around three inference axes: mechanism, control, and outcome.
    \item  We systematically evaluate a broad range of MLLMs, showing that current models struggle with zero-shot physical-field inference. We further explore model learning, transfer, and failure modes through domain-specific post-training and structured self-explanation analyses.
\end{itemize}
\section{Related Work}
\paragraph{Multimodal benchmarks for physics and scientific reasoning.} Recent multimodal benchmarks have increasingly evaluated whether vision-language models can reason about the physical world. Existing efforts can be broadly grouped into two lines. The first focuses on understanding common and intuitive physical phenomena in real-world environments, covering object properties, spatial relations, motion, and interaction dynamics, as in PhysBench~\citep{chow2025physbench} and PhysicsMind~\citep{mak2026physicsmind}. The second evaluates scientific reasoning through textbook-style physics problems involving physical laws, mathematical equations, diagrams, and numerical conditions, including SeePhys~\citep{xiang2026seephys}, PhyX~\citep{shen2025phyx}, and PhysUniBench~\citep{wang2025physunibench}. More broadly, multimodal benchmarks such as ScienceQA~\citep{lu2022learn}, MMMU~\citep{yue2024mmmu}, MathVista~\citep{lu2024mathvista}, and We-Math~\citep{qiao2025we} further assess science and mathematics problem solving across diverse visual formats. Together, these benchmarks have substantially advanced the evaluation of intuitive physical reasoning and multimodal scientific problem solving. However, they provide limited evaluation of a capability important for scientific and engineering agents: extracting physically meaningful information from continuous fields, where relevant evidence is encoded in field morphology, spatial variation, and temporal evolution. PhysFieldBench complements these efforts by evaluating physical inference directly from visualized field observations.
\vspace{-10pt}
\paragraph{Physical-field datasets for scientific machine learning.}
In parallel, the scientific machine learning community has developed a rich ecosystem of datasets and benchmarks built around physical fields \citep{wu2025propinn}. The large-scale training data constructed for PDEFormer~\citep{ye2024pdeformer} span a broad, systematically generated family of one-dimensional PDEs, with controlled variations in equation terms, coefficients, source functions, and initial conditions. Such parameterized equation data enable controlled comparisons of how different equation components and physical conditions shape field evolution. PDEBench~\citep{takamoto2022pdebench} provides time-dependent partial differential equation simulations for evaluating learned solvers and forecasting models. Large-scale simulation collections such as The Well~\citep{ohana2024well} cover diverse spatiotemporal systems, including fluid dynamics, acoustics, and magnetohydrodynamics. RealPDEBench~\citep{hu2026realpdebench}  further combines real-world measurements with numerical simulations to study scientific machine learning in real-world and sim-to-real settings. Physical-field data also arise in observational and engineering applications: SEVIR~\citep{veillette2020sevir} and TCIR~\citep{chen2018rotation} support weather forecasting and tropical-cyclone analysis, while DrivAerNet++~\citep{elrefaie2024drivaernet++}, DrivAerML~\citep{ashton2024drivaerml}, and NASA-CRM~\citep{bekemeyer2025introduction} provide high-fidelity computational fluid dynamics (CFD) data for aerodynamic analysis, design optimization, and surrogate modeling. These resources contain the kinds of spatial and temporal fields that motivate our benchmark, but they are primarily designed for numerical objectives: forecasting future states, accelerating simulations, or bridging simulation and reality. They therefore do not directly evaluate whether MLLMs can semantically interpret visualized physical fields and use the encoded information for physical understanding.
\vspace{-5pt}
\paragraph{Post-training for multimodal reasoning.}
Post-training has become central to adapting pretrained language and multimodal models for instruction following, specialized tasks, and reasoning~\citep{yin2024survey}. Unlike pre-training, which primarily establishes general representations and knowledge, post-training shapes model behavior through either supervised learning from task-level demonstrations, including final answers and explicit reasoning traces, or reinforcement learning from verifiable rewards~\citep{zelikman2022star,deepseekai2025r1}. Building on these advances, recent work has extended both paradigms to multimodal reasoning. Supervised approaches such as T-SciQ~\citep{wang2024t}, LLaVA-CoT~\citep{xu2025llava}, and MAmmoTH-VL~\citep{guo2025mammoth} use teacher-generated rationales, structured reasoning traces, or large-scale rationale-rich instruction data to improve multimodal reasoning. Reward-based methods such as MM-Eureka~\citep{meng2025mm} apply rule-based reinforcement learning to multidisciplinary multimodal reasoning, while Reason-RFT~\citep{tan2025reasonrft} combines chain-of-thought supervision with reinforcement learning to improve generalization across visual reasoning tasks. These studies show that post-training can improve performance on targeted multimodal reasoning capabilities, with some evidence of cross-task or cross-domain transfer. However, whether these findings extend to physical-field inference remains underexplored. In particular, it remains unclear how different post-training signals affect performance on training-covered physical-field tasks and whether the resulting gains generalize to held-out tasks. Using PhysFieldBench, we investigate these questions by comparing final-answer supervision, chain-of-thought supervision, and reinforcement learning under a unified evaluation protocol.
\section{PhysFieldBench}
To systematically evaluate whether MLLMs can extract and use physical information from continuous fields, we introduce PhysFieldBench. We define a task space spanning diverse field domains and inference targets, and construct a unified evaluation benchmark from heterogeneous physical-field data. We further establish a task-level split to assess cross-task generalization after post-training.
\subsection{Overview}
\begin{figure*}[t]
\centering
\includegraphics[width=\columnwidth]{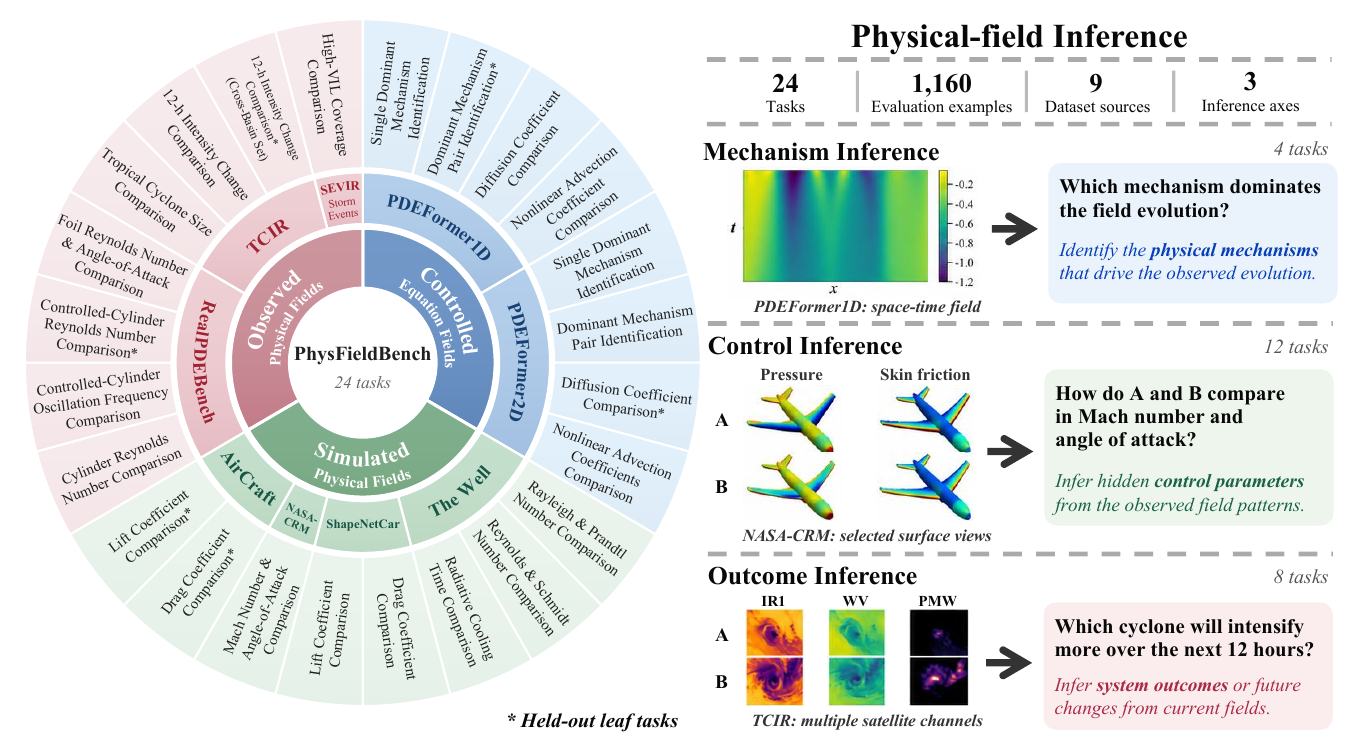}
\caption{Overview of PhysFieldBench. It comprises 24 tasks and 1,160 examples from nine sources, spanning controlled, simulated, and observed physical fields and three inference types: mechanism, control, and outcome. Asterisks denote the six held-out leaf tasks used to assess cross-task transfer.}
\label{fig:benchmark-overview}
\vspace{-15pt}
\end{figure*}

PhysFieldBench comprises 24 leaf tasks and 1,160 examples from nine datasets, as summarized in Figure \ref{fig:benchmark-overview}. Its task space is organized along two complementary dimensions: field domains characterize the origins of the observations, while inference axes specify the physical
information to be inferred.
\vspace{-5pt}
\paragraph{Task formulation.} 
We operationalize physical-field understanding through \emph{physical-field inference}, whose goal is to infer a physically grounded target from one or more field-structured observations. As illustrated in Figure~\ref{fig:benchmark-overview}, each benchmark example is represented as $(\mathcal{X}, q, \mathcal{C}, y)$, where $\mathcal{X}$ contains one or more visualized physical fields, $q$ is a task-specific question, $\mathcal{C}$ is the set of candidate answers, and $y \in \mathcal{C}$ is the correct answer. The physical information required to determine $y$ is not explicitly provided in the input and must instead be inferred from spatial structures, temporal evolution, field magnitudes, or relationships between observations. We define a leaf task as the finest-grained evaluation unit in PhysFieldBench, specified by a source dataset or subset, a physical inference target, and an example-construction protocol. This protocol determines how field observations are selected, represented, paired when applicable, and instantiated as a visual question-answering problem.
\vspace{-5pt}
\paragraph{Field domains.}
We organize the nine source datasets into three field domains according to how their fields are obtained. \emph{Controlled equation fields} are generated from parameterized PDE with systematic variations in governing terms and physical coefficients. This controllability links changes in field evolution to explicit mechanism and parameter settings, enabling grounded mechanism and control inference tasks. \emph{Simulated physical fields} are obtained from scientific simulations and engineering CFD, capturing complex geometries, boundary conditions, and coupled physical processes. \emph{Observed physical fields} comprise experimental measurements and meteorological observations, and therefore exhibit measurement noise, partial observability, and naturally occurring variation. Each domain contributes eight leaf tasks, providing balanced task-level coverage across three settings.
\vspace{-5pt}
\paragraph{Inference axes.}
Complementary to the field-domain taxonomy, we organize the 24 leaf tasks along three inference axes according to the physical target queried by each task. The benchmark contains 4 mechanism-inference tasks, 12 control-inference tasks, and 8 outcome-inference tasks. \emph{Mechanism inference} identifies the governing process or combination of processes responsible for the observed field evolution. \emph{Control inference} compares latent physical parameters or operating conditions that shape the observed field patterns. \emph{Outcome inference} compares system-level properties or future changes associated with the observed field states. Moreover, these axes describe complementary inference targets rather than successive levels of task difficulty. Together with the field domains, these axes organize PhysFieldBench along two complementary dimensions: how the field observations are obtained and what physical information must be inferred.
\begin{figure*}[t]
\centering
\includegraphics[width=\columnwidth]{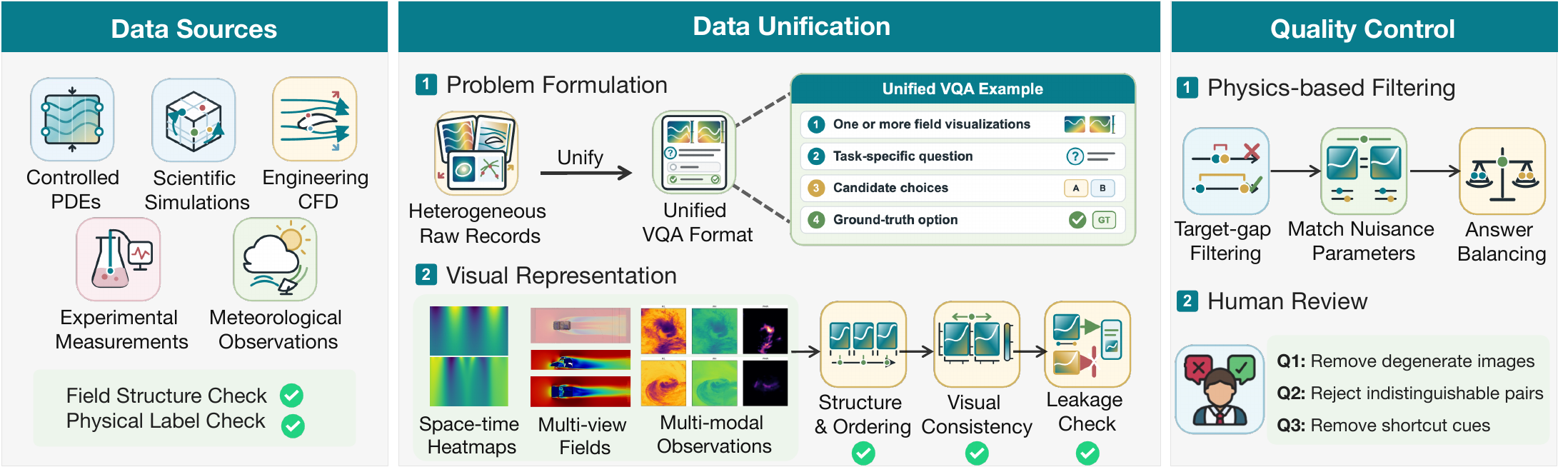}
\caption{{Construction pipeline of PhysFieldBench.}
Heterogeneous physical-field data first undergo physical-label verification, then are converted into a unified visual question-answering format with field-appropriate representations and further refined by physics-based filtering and human review.}
\label{fig:construction}
\vspace{-15pt}
\end{figure*}
\subsection{Benchmark Construction}
To instantiate the task space above, we construct PhysFieldBench through source selection, question construction, field rendering, and quality control, as illustrated in Figure~\ref{fig:construction}. This pipeline converts heterogeneous physical fields into a unified multiple-choice visual question-answering format, with answers grounded in physical records and visual inputs checked for ambiguity and answer leakage.
\vspace{-5pt}
\paragraph{Source selection.} We select source datasets that provide spatial or temporal field observations and physical records from which ground-truth answers can be derived. Mechanism labels are grounded in the governing equations, control-variable labels in recorded physical parameters or operating conditions, and outcome labels in simulation outputs or observational annotations. These criteria yield nine source datasets across the three field domains, providing the field observations and target labels used to construct benchmark examples.
\vspace{-5pt}
\paragraph{Question construction.}
We formulate each physical inference target as a multiple-choice question. For categorical targets, such as a dominant mechanism or a combination of mechanisms, candidate answers correspond to the possible categories. For continuous targets, we construct pairwise comparisons in which the model determines the relative ordering of one or more physical quantities between two field observations. The correct option is determined from the corresponding ground-truth labels or values. This comparison format avoids direct regression across quantities with different physical units and scales while retaining the task of inferring their ordering from visual field evidence.
\vspace{-5pt}
\paragraph{Field rendering.}
We render physical fields according to their spatial and temporal structure: time-dependent fields as space-time heatmaps or frame montages, CFD fields as multi-view or multi-variable panels, and meteorological observations as multimodal panels. Within each input image, all temporal frames or views of the same physical quantity share a single colorbar. We retain calibrated colorbars because field magnitudes provide physical evidence for inference, while removing textual labels, parameter values, and case identifiers that directly reveal the answer.
\vspace{-5pt}
\paragraph{Quality control.}
We enforce disjointness across leaf tasks at the level of underlying physical cases, rather than rendered images or question instances. For pairwise tasks, we apply task-specific thresholds on the difference in ground-truth target values to exclude near-tie pairs. Where metadata permit, we also match or constrain non-target conditions to reduce alternative cues for the comparison. We balance answer positions and manually review all evaluation examples, removing corrupted renderings, visually indistinguishable pairs, inconsistent labels, and residual answer-revealing cues. Together, these checks reduce ambiguity and direct shortcuts while retaining physically valid variation.
\vspace{-5pt}
\paragraph{Overall benchmark.}
The final evaluation set contains 1,160 examples across 24 leaf tasks drawn from nine source datasets. By combining diverse field domains with complementary inference targets, PhysFieldBench evaluates physical-field understanding across a broad range of physical systems and observation settings. The construction pipeline supports this breadth with physically grounded answers, consistent field rendering within each task, and systematic quality control to reduce ambiguous comparisons and answer leakage. Together, these design choices establish a broad and carefully curated testbed for evaluating how well MLLMs infer physical information from fields. 
To assess cross-task transfer, we designate 18 tasks as training-covered and hold out the remaining six from post-training. The held-out set includes two tasks from each field domain and collectively covers mechanism, control, and outcome inference. Although covered and held-out tasks may draw from the same source dataset, their underlying physical cases remain disjoint. This design evaluates whether models can recombine related atomic abilities learned from training-covered tasks to solve tasks with physical targets or observation settings not covered during post-training.

\section{Experiments}

Built upon PhysFieldBench, we address three questions:  (1) How well do current MLLMs perform physical-field inference? (2) Where do they fail in interpreting field observations?~(3) Can post-training improve performance and cross-task generalization? We investigate these questions through zero-shot evaluation with a supervised visual baseline, structured self-explanation analysis, and comparisons of answer-only supervision, chain-of-thought supervision, and reinforcement learning.
\vspace{-5pt}
\paragraph{Evaluated models.}
We evaluate open-source MLLMs, including Qwen3-VL-8B-Instruct~\citep{bai2025qwen3vl}, Intern-S1-Pro~\citep{zou2026interns1pro}, LLaVA-OneVision-2-8B~\citep{an2026llavaonevision2}, and Kimi K2.6~\citep{moonshot2026kimik26}, alongside proprietary models, including GPT-5.5~\citep{openai2026gpt55}, Gemini 3 Flash~\citep{deepmind2025gemini3flash}, Gemini 3.1 Pro~\citep{deepmind2026gemini31pro}, Gemini 3.5 Flash~\citep{deepmind2026gemini35flash}, Claude Opus 4.7~\citep{anthropic2026claudeopus47}, Qwen3.6-Plus~\citep{qwen2026qwen36plus}, Doubao-Seed-2.0-Pro~\citep{bytedance2026seed20}, and GLM-5V-Turbo~\citep{vteam2026glm5vturbo}. We additionally train a lightweight Vision Transformer (ViT)~\citep{dosovitskiy2021vit} separately for each task. Each ViT is trained on a task-specific training set that shares the distribution of the corresponding test set but contains none of its samples, and is then evaluated on PhysFieldBench. This baseline is not directly comparable to zero-shot MLLMs, but provides a supervised reference for assessing whether the visualized physical fields contain learnable physical signals.
\vspace{-5pt}
\paragraph{Evaluation protocol.}
We evaluate all MLLMs in a multiple-choice visual question-answering format, providing the field visualizations, question, and answer choices for each example. Unless otherwise specified, the main benchmark comparison uses zero-shot prompting without in-context examples. We extract the final answer choice from each model response and compare it with the ground truth. Responses that cannot be parsed into a valid answer choice are counted as incorrect.
\vspace{-5pt}
\paragraph{Evaluation metrics.}
We report raw accuracy and a chance-normalized score to account for differences in the number of answer choices across tasks. For each task, we compute $\mathrm{NormScore} = 100 \times \frac{a-r}{1-r}$, where $a$ is the accuracy and $r=1/K$ is the chance accuracy under uniform guessing over $K$ answer choices. A score of $0$ indicates chance-level performance, $100$ indicates perfect accuracy, and negative scores indicate performance below chance. We report macro-averaged scores over all 24 tasks, as well as separately for each field domain and inference axis.
\begin{table*}[t]
\begin{center}
\setlength{\tabcolsep}{1.3pt}
\begin{minipage}{\textwidth}
\caption{
Main results on PhysFieldBench. We report chance-normalized scores, macro-averaged over leaf tasks within each field domain and inference axis. Overall reports the macro-average over all tasks. A score of 0 indicates chance-level performance. MLLMs are evaluated zero-shot, while ViT is trained separately for each task on i.i.d. training data drawn from the same distribution as the test set. Bold and underlined values denote the best and second-best scores among MLLMs, respectively.
}
\vspace{-8pt}
\label{tab:main_results}
\vskip 3pt
\begin{small}
\begin{tabular*}{\textwidth}{@{\extracolsep{\fill}}l|ccc|ccc|cc@{}}
\toprule
\multirow{2}{*}{Models}
& \multicolumn{3}{c|}{Field Domains}
& \multicolumn{3}{c|}{Inference Axes}
& \multirow{2}{*}{Overall}
& \multirow{2}{*}{Rank} \\
\cmidrule(lr){2-4} \cmidrule(lr){5-7}
& Controlled & Simulated & Observed
& Mechanism & Control & Outcome
& & \\
\midrule
\rowcolor{groupgray}
\multicolumn{9}{@{}l}{\textit{Supervised Visual Reference}} \\
ViT~(\citeyear{dosovitskiy2021vit})
& 54.8 & 76.3 & 80.1
& 63.6 & 73.3 & 69.5
& 70.4 & - \\
\midrule
\rowcolor{groupgray}
\multicolumn{9}{@{}l}{\textit{Open-source Models}} \\
Qwen3-VL-8B-Instruct~(\citeyear{bai2025qwen3vl})
& -1.3 & -4.3 & 13.6
& 6.0 & -3.0 & 9.5
& 2.7 & 11 \\
LLaVA-OneVision-2-8B~(\citeyear{an2026llavaonevision2})
& 0.1 & -7.4 & 1.5
& 2.2 & -1.2 & -5.0
& -1.9 & 12 \\
Intern-S1-Pro~(\citeyear{zou2026interns1pro})
& 14.3 & 12.9 & 14.0
& 15.1 & 5.4 & 25.5
& 13.7 & 9 \\
Kimi K2.6~(\citeyear{moonshot2026kimik26})
& 30.9 & 22.5 & 18.1
& 38.3 & 14.3 & 31.0
& 23.9 & 6 \\
\midrule
\rowcolor{groupgray}
\multicolumn{9}{@{}l}{\textit{Proprietary Models}} \\
GLM-5V-Turbo~(\citeyear{vteam2026glm5vturbo})
& 23.1 & 11.7 & \textbf{22.8}
& 25.2 & 12.6 & 26.0
& 19.2 & 8 \\
Qwen3.6-Plus~(\citeyear{qwen2026qwen36plus})
& 33.4 & 22.9 & 21.5
& 35.3 & 19.7 & 30.5
& \underline{25.9} & 2 \\
Doubao-Seed-2.0-Pro~(\citeyear{bytedance2026seed20})
& 24.1 & \textbf{30.9} & 14.9
& 32.0 & \underline{19.9} & 24.0
& 23.3 & 7 \\
Claude Opus 4.7~(\citeyear{anthropic2026claudeopus47})
& 24.3 & 1.0 & -4.5
& 46.5 & 1.4 & -4.5
& 6.9 & 10 \\
Gemini 3 Flash~(\citeyear{deepmind2025gemini3flash})
& 31.0 & \underline{30.6} & 16.2
& 37.3 & 16.7 & 34.0
& \underline{25.9} & 2 \\
Gemini 3.1 Pro~(\citeyear{deepmind2026gemini31pro})
& 38.0 & 22.5 & 16.6
& \underline{46.8} & 9.1 & \textbf{40.0}
& 25.7 & 4 \\
Gemini 3.5 Flash~(\citeyear{deepmind2026gemini35flash})
& \underline{41.6} & 23.5 & 8.9
& 46.5 & 9.2 & \underline{37.0}
& 24.7 & 5 \\
GPT-5.5~(\citeyear{openai2026gpt55})
& \textbf{43.5} & 21.6 & \underline{22.7}
& \textbf{48.1} & \textbf{22.5} & 30.0
& \textbf{29.3} & 1 \\
\bottomrule
\end{tabular*}
\end{small}
\end{minipage}
\end{center}
\vspace{-18pt}
\end{table*}
\subsection{Main Results}
\vspace{-2pt}
\paragraph{Overall performance.}
Table~\ref{tab:main_results} shows that current MLLMs achieve limited zero-shot performance on PhysFieldBench. GPT-5.5 obtains the highest overall normalized score of 29.3, while all other MLLMs score below 26. Gemini 3 Flash, Gemini 3.1 Pro, and Gemini 3.5 Flash achieve similar scores of 25.9, 25.7, and 24.7, respectively, suggesting that general model iteration does not necessarily translate into consistent gains in physical-field inference. Qwen3.6-Plus matches Gemini 3 Flash at 25.9, while Kimi K2.6 and Doubao-Seed-2.0-Pro are also competitive at 23.9 and 23.3, followed by GLM-5V-Turbo at 19.2. Claude Opus 4.7 scores only 6.9 overall despite strong mechanism-inference performance, indicating uneven capabilities across tasks. The two 8B models, Qwen3-VL-8B-Instruct and LLaVA-OneVision-2-8B, remain near chance at 2.7 and -1.9, although their smaller scale may contribute to this limitation. In contrast, the supervised ViT reference reaches 70.4, showing that task-specific supervision enables a vision model to learn predictive signals from the field visualizations. Although the training settings differ, these results reveal a substantial gap between performance under task-specific supervision and the zero-shot physical-field inference capabilities of general-purpose MLLMs. Detailed inference settings for all evaluated models are provided
in Appendix~\ref{app:inference_settings}.
\vspace{-5pt}
\paragraph{Performance across field domains.}
Nine of the twelve MLLMs achieve their highest scores on controlled equation fields, indicating substantial variation in performance across field domains. GPT-5.5 scores 43.5 on controlled fields, compared with 21.6 on simulated fields and 22.7 on observed fields. Gemini 3.5 Flash exhibits an even wider spread, scoring 41.6, 23.5, and 8.9, respectively. However, the leading model also varies by domain. While GPT-5.5 leads on controlled fields, Doubao-Seed-2.0-Pro and Gemini 3 Flash rank first and second on simulated fields, scoring 30.9 and 30.6, respectively. On observed fields, GLM-5V-Turbo leads with 22.8 despite its lower overall score of 19.2. These differences highlight model strengths beyond the overall ranking.
\vspace{-5pt}
\paragraph{Performance across inference axes.}
Control inference is particularly challenging for current MLLMs, with ten of the twelve models achieving their lowest scores on this axis. GPT-5.5 leads on control inference with 22.5, yet this remains well below its mechanism and outcome scores of 48.1 and 30.0. The disparity is more pronounced for Gemini 3.1 Pro, which scores 46.8 on mechanism inference and 40.0 on outcome inference, but only 9.1 on control inference. Claude Opus 4.7 further illustrates this uneven performance: its mechanism score of 46.5 contrasts with 1.4 on control and -4.5 on outcome inference. These results suggest that recognizing governing mechanisms does not necessarily translate into reliable inference of latent conditions or system outcomes.
\begin{figure*}[t]
\centering
\includegraphics[width=\columnwidth]{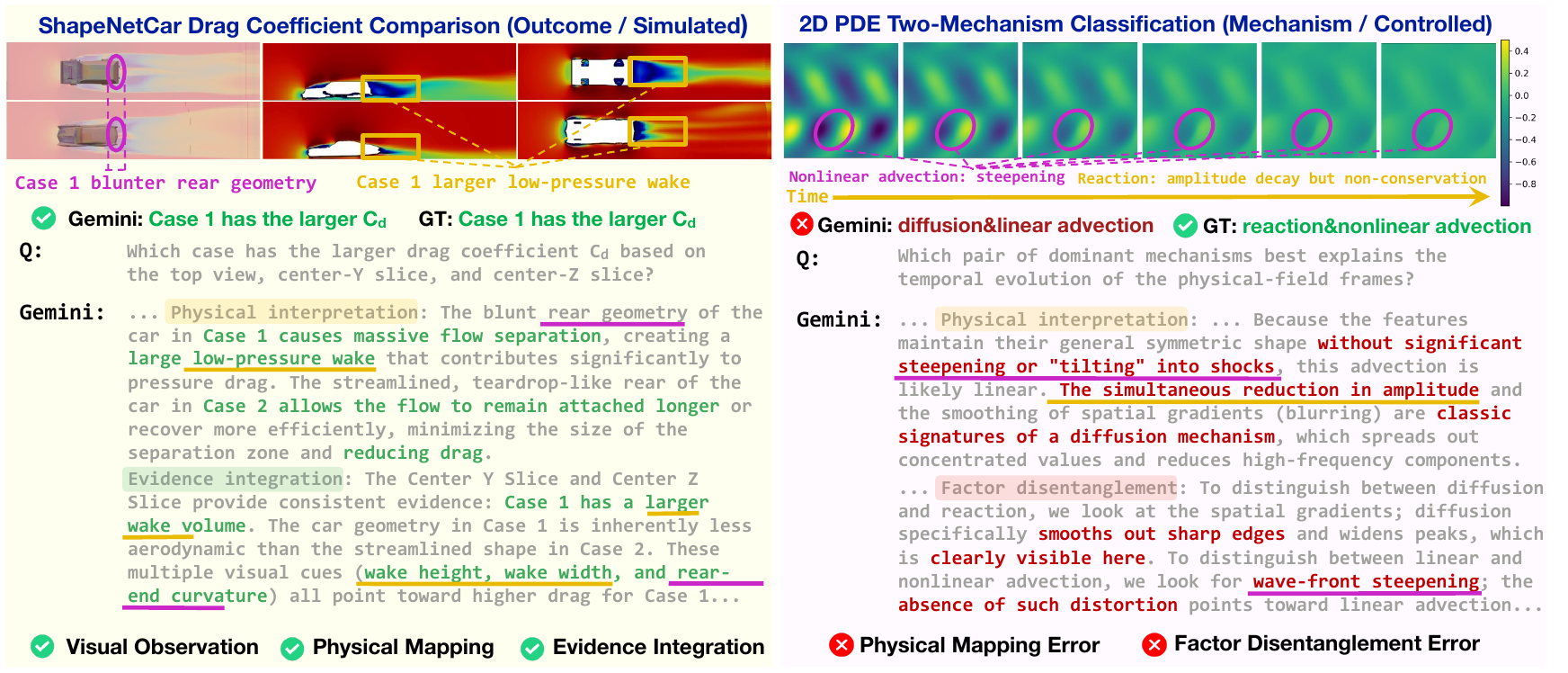}
\caption{Self-explanation showcases. Gemini 3 Flash correctly integrates geometric and multi-view evidence in ShapeNetCar but shows physical-mapping and factor-disentanglement errors in 2D PDE.}
\label{fig:case_study}
\vspace{5pt}
\includegraphics[width=\columnwidth]{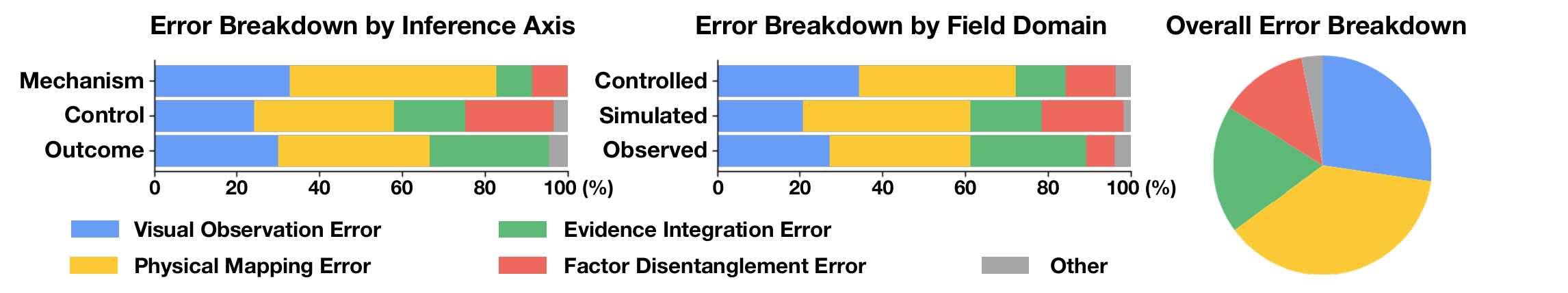}
\caption{Self-explanation error analysis. Error responses are summarized by inference axis and field domain, with visual-observation errors and physical-mapping errors being the most common.}
\label{fig:error_analysis}
\vspace{-15pt}
\end{figure*}
\vspace{-2pt}
\subsection{Probing via Self-Explanations}
\vspace{-2pt}
\paragraph{Setup.} To diagnose where MLLMs struggle in physical-field inference, we conduct a structured self-explanation analysis using Gemini 3 Flash.  We use these self-explanations as diagnostic probes of the model's explicit visual-to-physical reasoning. Specifically, we replace the original answer-only instruction with a structured rationale template that organizes the required inference into six diagnostic stages: visual observation, physical mapping, evidence integration, pairwise comparison, factor disentanglement, and target alignment. From this self-explanation run, we select 162 incorrectly answered examples across all 24 tasks using task-stratified sampling. Given the task input, the generated self-explanation and final answer, and the ground truth, GPT-5.5 labels each failure with one to three applicable error categories corresponding to these stages and ranks them by their contribution to the incorrect prediction. More details are provided in Appendix~\ref{appendix:self_explain}.
\vspace{-7pt}
\paragraph{Case studies.} 
Figure~\ref{fig:case_study} contrasts a successful and a failed self-explanation, illustrating how the structured probe exposes different stages of visual-to-physical inference. In the ShapeNetCar drag-coefficient task, Gemini correctly identifies Case 1 as having the larger drag coefficient. The explanation forms a coherent inference chain: it observes the blunter rear geometry and more pronounced wake of Case 1, maps these cues to stronger flow separation and drag, and integrates consistent evidence across multiple views to support the final comparison. The 2D PDE example, in contrast, illustrates a failure in physical mapping and factor disentanglement. Gemini identifies physically relevant changes in the field, including amplitude decay and spatial smoothing, but interprets these cues as signatures of diffusion and takes the lack of an isolated steepening pattern as evidence for linear advection. The ground truth instead contains reaction and nonlinear advection, whose effects jointly shape the observed evolution and are therefore not cleanly separable by such one-to-one visual heuristics. The model thus produces a locally plausible physical explanation while mapping the observed patterns to the wrong mechanisms. This contrast shows that fluent explanations and recognition of salient field patterns do not necessarily imply reliable visual-to-physical grounding.
\vspace{-9pt}
\paragraph{Error analysis.}
We analyze the 162 incorrect Gemini 3 Flash examples and categorize their failures according to the six-stage reasoning process defined above. Specifically, errors in \textcolor{visualblue}{visual observation}, \textcolor{mappingyellow}{physical mapping}, \textcolor{integrationgreen}{evidence integration}, \textcolor{disentanglered}{factor disentanglement} are retained as four corresponding categories, while the less frequent pairwise-comparison and target-alignment failures are merged into \textcolor{othergray}{other}. As shown in Figure~\ref{fig:error_analysis}, \textcolor{mappingyellow}{physical mapping errors} and \textcolor{visualblue}{visual observation errors} are the most common among all assigned error labels. Moreover, 89.5\% of incorrect examples involve multiple error types, indicating that failures often span several stages of the visual-to-physical inference chain. Error patterns also vary by task: mechanism inference is dominated by physical-mapping failures, control inference shows more factor-disentanglement errors, and observed fields exhibit more evidence-integration failures. Overall, the main bottleneck lies not only in recognizing field patterns, but in grounding and combining them into correct physical interpretations.
\subsection{Post-training}
\begin{figure*}[t]
\centering
\includegraphics[width=\columnwidth]{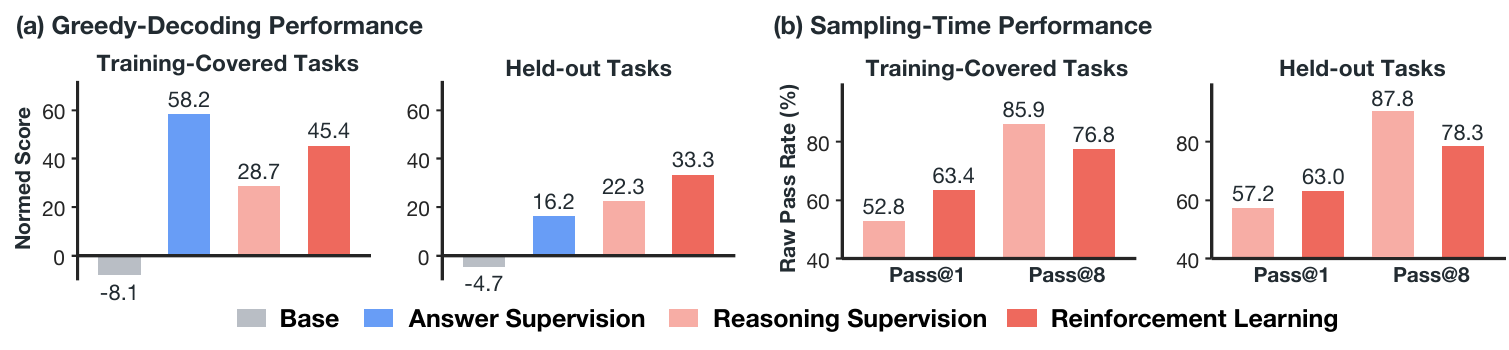}
\vspace{-15pt}
\caption{Post-training results. (a) Normed score on training-covered and held-out tasks. (b) Raw Pass@1 accuracy and Pass@8 coverage (\%) after reasoning supervision and reinforcement learning.}
\label{fig:post_training_results}
\vspace{-15pt}
\end{figure*}
To investigate whether domain-specific post-training improves physical-field inference and cross-task transfer, we compare three post-training strategies: final-answer supervision, structured reasoning supervision, and reinforcement learning with verifiable answer rewards. We first train the model using final answers only, then introduce teacher-generated reasoning traces, and finally apply reinforcement learning on top of the reasoning-supervised model. We report chance-normalized performance on training-covered and held-out tasks, together with Pass@1 and Pass@8 for sampling-time evaluation. Pass@8 counts an example as correct if any of eight sampled responses is correct.
\vspace{-5pt}
\paragraph{Answer supervision.}
We first fine-tune Qwen3-VL-8B-Thinking on the 18 training-covered tasks using 250 examples per task, with exactly the same training data as the supervised ViT reference. The model is supervised only with final answers, without explicit reasoning traces. As shown in Figure~\ref{fig:post_training_results}(a), answer supervision substantially improves the normalized score on training-covered tasks from $-8.1$ to $58.2$, but achieves only $16.2$ on held-out tasks, revealing a large gap between fitting the supervised tasks and transferring to unseen ones. The task split is detailed in Appendix~\ref{app:task_split}.
\vspace{-14pt}
\paragraph{Reasoning supervision.}
We then fine-tune Qwen3-VL-8B-Thinking with reasoning supervision. The chain of thought data are constructed from the same training pool used for answer supervision, but are substantially smaller. We use Qwen3.6 Plus to generate task-specific reasoning traces, prioritizing trajectories that independently reach the correct answer and supplementing insufficient cases with answer-conditioned generation. The retained traces are then rewritten by Gemini 3 Flash into the six-stage reasoning structure used in our self-explanation analysis, yielding 1,012 training examples. Figure~\ref{fig:post_training_results}(b) reveals substantial sampling-time potential: Pass@8 reaches $85.9\%$ on training-covered tasks and $87.8\%$ on held-out tasks. On training-covered tasks, the increase from Pass@1 ($52.8\%$) to Pass@8 ($85.9\%$) indicates that correct answers are often reachable under repeated sampling but are not consistently obtained in a single sample, suggesting substantial headroom for further RL.
\vspace{-7pt}
\paragraph{Reinforcement learning.}
Starting from the reasoning-supervised model, we further apply reinforcement learning with verifiable answer rewards. We train on 2,000 examples from the 18 training-covered tasks, excluding those used for reasoning supervision. As shown in Figure~\ref{fig:post_training_results}(a), reinforcement learning improves the normalized score from $28.7$ to $45.4$ on training-covered tasks and from $22.3$ to $33.3$ on held-out tasks. Figure~\ref{fig:post_training_results}(b) shows an accuracy--coverage trade-off after RL: Pass@1 improves substantially on both training-covered and held-out tasks, demonstrating that the gains generalize beyond the training-covered tasks, while Pass@8 coverage decreases. Further, this also suggests a narrower reasoning-capability boundary after RL, with more reliable dominant solutions, as intended, but lower coverage of correct reasoning trajectories \citep{yue2025reasoningcapacity}.

\section{Conclusion}
We introduce PhysFieldBench, a benchmark for evaluating physical-field understanding in multimodal models. It spans controlled equation fields, simulated physical fields, and observed physical fields, and requires models to infer governing mechanisms, latent control variables, and outcome properties directly from field-structured observations. Across these diverse settings, current MLLMs remain far from reliable physical-field inference despite the presence of learnable visual signals. Our diagnostic analyses further show that failures arise from missed visual evidence and unreliable mappings between field patterns and their physical meanings. Post-training substantially improves task performance, yet the gap between covered and held-out tasks indicates cross-task generalization remains challenging.
\newpage
\subsection*{AI use statement}
This work evaluates multimodal large language models (MLLMs) as research subjects. We also used generative AI tools to support several distinct parts of the research process: Qwen3.6-Plus generated candidate reasoning traces, Gemini 3 Flash structured these traces and produced self-explanations for diagnostic analysis, and GPT-5.5 assisted with error categorization. ChatGPT assisted with English-language editing of the manuscript. The authors reviewed the AI-assisted materials used in this work and take responsibility for the final benchmark data, analyses, and manuscript.
\bibliography{iclr2027_conference}

@inproceedings{chow2025physbench,
  title={PhysBench: Benchmarking and Enhancing Vision-Language Models for Physical World Understanding},
  author={Chow, Wei and Mao, Jiageng and Li, Boyi and Seita, Daniel and Guizilini, Vitor and Wang, Yue},
  booktitle={International Conference on Learning Representations},
  year={2025}
}

@article{mak2026physicsmind,
  title={PhysicsMind: Sim and Real Mechanics Benchmarking for Physical Reasoning and Prediction in Foundational VLMs and World Models},
  author={Mak, Chak-Wing and Zhu, Guanyu and Zhang, Boyi and Li, Hongji and Chi, Xiaowei and Zhang, Kevin and Wu, Yichen and He, Yangfan and Fan, Chun-Kai and Lu, Wentao and others},
  journal={arXiv preprint arXiv:2601.16007},
  year={2026}
}

@article{xiang2026seephys,
  title={Seephys: Does seeing help thinking?--benchmarking vision-based physics reasoning},
  author={Xiang, Kun and Li, Heng and Zhang, Terry Jingchen and Huang, Yinya and Liu, Zirong and Qu, Peixin and He, Jixi and Chen, Jiaqi and Yuan, Yu-Jie and Han, Jianhua and others},
  journal={Advances in Neural Information Processing Systems},
  volume={38},
  year={2026}
}

@article{shen2025phyx,
  title={PhyX: Does Your Model Have the" Wits" for Physical Reasoning?},
  author={Shen, Hui and Wu, Taiqiang and Han, Qi and Hsieh, Yunta and Wang, Jizhou and Zhang, Yuyue and Cheng, Yuxin and Hao, Zijian and Ni, Yuansheng and Wang, Xin and others},
  journal={arXiv preprint arXiv:2505.15929},
  year={2025}
}

@article{wang2025physunibench,
  title={Physunibench: an undergraduate-level physics reasoning benchmark for multimodal models},
  author={Wang, Lintao and Su, Encheng and Liu, Jiaqi and Li, Pengze and Xia, Peng and Xiao, Jiabei and Zhang, Wenlong and Dai, Xinnan and Chen, Xi and Meng, Yuan and others},
  journal={arXiv preprint arXiv:2506.17667},
  year={2025}
}

@article{lu2022learn,
  title={Learn to explain: Multimodal reasoning via thought chains for science question answering},
  author={Lu, Pan and Mishra, Swaroop and Xia, Tanglin and Qiu, Liang and Chang, Kai-Wei and Zhu, Song-Chun and Tafjord, Oyvind and Clark, Peter and Kalyan, Ashwin},
  journal={Advances in neural information processing systems},
  volume={35},
  pages={2507--2521},
  year={2022}
}

@inproceedings{yue2024mmmu,
  title={Mmmu: A massive multi-discipline multimodal understanding and reasoning benchmark for expert agi},
  author={Yue, Xiang and Ni, Yuansheng and Zhang, Kai and Zheng, Tianyu and Liu, Ruoqi and Zhang, Ge and Stevens, Samuel and Jiang, Dongfu and Ren, Weiming and Sun, Yuxuan and others},
  booktitle={Proceedings of the IEEE/CVF conference on computer vision and pattern recognition},
  pages={9556--9567},
  year={2024}
}

@inproceedings{lu2024mathvista,
  title={Mathvista: Evaluating mathematical reasoning of foundation models in visual contexts},
  author={Lu, Pan and Bansal, Hritik and Xia, Tony and Liu, Jiacheng and Li, Chunyuan and Hajishirzi, Hannaneh and Cheng, Hao and Chang, Kai-Wei and Galley, Michel and Gao, Jianfeng},
  booktitle={International Conference on Learning Representations},
  volume={2024},
  pages={23439--23554},
  year={2024}
}

@inproceedings{qiao2025we,
  title={We-math: Does your large multimodal model achieve human-like mathematical reasoning?},
  author={Qiao, Runqi and Tan, Qiuna and Dong, Guanting and MinhuiWu, MinhuiWu and Sun, Chong and Song, Xiaoshuai and Wang, Jiapeng and Gongque, Zhuoma and Lei, Shanglin and Zhang, Yifan and others},
  booktitle={Proceedings of the 63rd Annual Meeting of the Association for Computational Linguistics (Volume 1: Long Papers)},
  pages={20023--20070},
  year={2025}
}

@inproceedings{takamoto2022pdebench,
  title={PDEBench: An Extensive Benchmark for Scientific Machine Learning},
  author={Takamoto, Makoto and Praditia, Timothy and Leiteritz, Raphael and MacKinlay, Dan and Alesiani, Francesco and Pfl{\"u}ger, Dirk and Niepert, Mathias},
  booktitle={Advances in Neural Information Processing Systems Datasets and Benchmarks Track},
  year={2022}
}

@article{ye2024pdeformer,
  title={Pdeformer: Towards a foundation model for one-dimensional partial differential equations},
  author={Ye, Zhanhong and Huang, Xiang and Chen, Leheng and Liu, Hongsheng and Wang, Zidong and Dong, Bin},
  journal={arXiv preprint arXiv:2402.12652},
  year={2024}
}

@inproceedings{ohana2024well,
  title={The Well: a Large-Scale Collection of Diverse Physics Simulations for Machine Learning},
  author={Ohana, Ruben and McCabe, Michael and Meyer, Lucas and Morel, Rudy and Agocs, Fruzsina J. and Beneitez, Miguel and Berger, Marsha and Burkhart, Blakesley and Burns, Keaton and Dalziel, Stuart B. and others},
  booktitle={Advances in Neural Information Processing Systems Datasets and Benchmarks Track},
  year={2024}
}

@inproceedings{hu2026realpdebench,
  title={RealPDEBench: A Benchmark for Complex Physical Systems with Real-World Data},
  author={Hu, Peiyan and Feng, Haodong and Liu, Hongyuan and Yan, Tongtong and Deng, Wenhao and Gao, Tianrun and Zheng, Rong and Zheng, Haoren and Yu, Chenglei and Wang, Chuanrui and Li, Kaiwen and Ma, Zhi-Ming and Zhou, Dezhi and Lu, Xingcai and Fan, Dixia and Wu, Tailin},
  booktitle={International Conference on Learning Representations},
  year={2026}
}

@article{veillette2020sevir,
  title={Sevir: A storm event imagery dataset for deep learning applications in radar and satellite meteorology},
  author={Veillette, Mark and Samsi, Siddharth and Mattioli, Chris},
  journal={Advances in neural information processing systems},
  volume={33},
  pages={22009--22019},
  year={2020}
}

@inproceedings{chen2018rotation,
  title={Rotation-blended CNNs on a new open dataset for tropical cyclone image-to-intensity regression},
  author={Chen, Boyo and Chen, Buo-Fu and Lin, Hsuan-Tien},
  booktitle={Proceedings of the 24th ACM SIGKDD international conference on knowledge discovery \& data mining},
  pages={90--99},
  year={2018}
}

@article{elrefaie2024drivaernet++,
  title={Drivaernet++: A large-scale multimodal car dataset with computational fluid dynamics simulations and deep learning benchmarks},
  author={Elrefaie, Mohamed and Morar, Florin and Dai, Angela and Ahmed, Faez},
  journal={Advances in Neural Information Processing Systems},
  volume={37},
  pages={499--536},
  year={2024}
}

@article{ashton2024drivaerml,
  title={DrivAerML: High-fidelity computational fluid dynamics dataset for road-car external aerodynamics},
  author={Ashton, Neil and Mockett, Charles and Fuchs, Marian and Fliessbach, Louis and Hetmann, Hendrik and Knacke, Thilo and Schonwald, Norbert and Skaperdas, Vangelis and Fotiadis, Grigoris and Walle, Astrid and others},
  journal={arXiv preprint arXiv:2408.11969},
  year={2024}
}

@inproceedings{bekemeyer2025introduction,
  title={Introduction of applied aerodynamics surrogate modeling benchmark cases},
  author={Bekemeyer, Philipp and Hariharan, Nathan and Wissink, Andrew M and Cornelius, Jason},
  booktitle={AIAA SCITECH 2025 Forum},
  pages={0036},
  year={2025}
}

@inproceedings{wang2024t,
  title={T-sciq: Teaching multimodal chain-of-thought reasoning via large language model signals for science question answering},
  author={Wang, Lei and Hu, Yi and He, Jiabang and Xu, Xing and Liu, Ning and Liu, Hui and Shen, Heng Tao},
  booktitle={Proceedings of the AAAI Conference on Artificial Intelligence},
  volume={38},
  number={17},
  pages={19162--19170},
  year={2024}
}

@inproceedings{xu2025llava,
  title={Llava-cot: Let vision language models reason step-by-step},
  author={Xu, Guowei and Jin, Peng and Wu, Ziang and Li, Hao and Song, Yibing and Sun, Lichao and Yuan, Li},
  booktitle={Proceedings of the IEEE/CVF International Conference on Computer Vision},
  pages={2087--2098},
  year={2025}
}

@inproceedings{guo2025mammoth,
  title={Mammoth-vl: Eliciting multimodal reasoning with instruction tuning at scale},
  author={Guo, Jiawei and Zheng, Tianyu and Li, Yizhi and Bai, Yuelin and Li, Bo and Wang, Yubo and Zhu, King and Neubig, Graham and Chen, Wenhu and Yue, Xiang},
  booktitle={Proceedings of the 63rd Annual Meeting of the Association for Computational Linguistics (Volume 1: Long Papers)},
  pages={13869--13920},
  year={2025}
}

@misc{meng2025mm,
  title={MM-Eureka: Exploring the Frontiers of Multimodal Reasoning with Rule-based Reinforcement Learning},
  author={Meng, Fanqing and Du, Lingxiao and Liu, Zongkai and Zhou, Zhixiang and Lu, Quanfeng and Fu, Daocheng and Han, Tiancheng and Shi, Botian and Wang, Wenhai and He, Junjun and Zhang, Kaipeng and Luo, Ping and Qiao, Yu and Zhang, Qiaosheng and Shao, Wenqi},
  year={2025},
  eprint={2503.07365},
  archivePrefix={arXiv},
  primaryClass={cs.CV}
}

@inproceedings{liu2023visualinstruction,
  title     = {Visual Instruction Tuning},
  author    = {Liu, Haotian and Li, Chunyuan and Wu, Qingyang and Lee, Yong Jae},
  booktitle = {Advances in Neural Information Processing Systems},
  volume    = {36},
  pages     = {34892--34916},
  year      = {2023},
  doi       = {10.52202/075280-1516}
}

@article{zhou2024unisolver,
  title={Unisolver: Pde-conditional transformers towards universal neural pde solvers},
  author={Zhou, Hang and Ma, Yuezhou and Wu, Haixu and Wang, Haowen and Long, Mingsheng},
  journal={arXiv preprint arXiv:2405.17527},
  year={2024}
}

@inproceedings{NEURIPS2025_332b4fbe,
  title={PhySense: Sensor Placement Optimization for Accurate Physics Sensing},
  author={Ma, Yuezhou and Wu, Haixu and Zhou, Hang and Weng, Huikun and Wang, Jianmin and Long, Mingsheng},
  booktitle={Advances in Neural Information Processing Systems},
  volume={38},
  pages={35697--35725},
  year={2025},
  publisher={Curran Associates, Inc.},
  doi={10.52202/085713-1200}
}

@article{comanici2025gemini25,
  title   = {Gemini 2.5: Pushing the Frontier with Advanced Reasoning,
             Multimodality, Long Context, and Next Generation Agentic Capabilities},
  author  = {Comanici, Gheorghe and others},
  journal = {arXiv preprint arXiv:2507.06261},
  year    = {2025},
  doi     = {10.48550/arXiv.2507.06261}
}

@article{openai2023gpt4,
  title   = {{GPT}-4 Technical Report},
  author  = {{OpenAI}},
  journal = {arXiv preprint arXiv:2303.08774},
  year    = {2023},
  url     = {https://arxiv.org/abs/2303.08774}
}

@article{wang2023scientificdiscovery,
  title   = {Scientific Discovery in the Age of Artificial Intelligence},
  author  = {Wang, Hanchen and Fu, Tianfan and Du, Yuanqi and Gao, Wenhao and
             Huang, Kexin and Liu, Ziming and Chandak, Payal and Liu, Shengchao and
             Van Katwyk, Peter and Deac, Andreea and Anandkumar, Anima and
             Bergen, Karianne and Gomes, Carla P. and Ho, Shirley and Kohli, Pushmeet and
             Lasenby, Joan and Leskovec, Jure and Liu, Tie-Yan and Manrai, Arjun and
             Marks, Debora and Ramsundar, Bharath and Song, Le and Sun, Jimeng and
             Tang, Jian and Veli{\v{c}}kovi{\'c}, Petar and Welling, Max and
             Zhang, Linfeng and Coley, Connor W. and Bengio, Yoshua and Zitnik, Marinka},
  journal = {Nature},
  volume  = {620},
  number  = {7972},
  pages   = {47--60},
  year    = {2023},
  doi     = {10.1038/s41586-023-06221-2}
}

@article{boiko2023autonomous,
  title   = {Autonomous Chemical Research with Large Language Models},
  author  = {Boiko, Daniil A. and MacKnight, Robert and Kline, Ben and Gomes, Gabe},
  journal = {Nature},
  volume  = {624},
  number  = {7992},
  pages   = {570--578},
  year    = {2023},
  doi     = {10.1038/s41586-023-06792-0}
}

@article{bran2024augmenting,
  title   = {Augmenting Large Language Models with Chemistry Tools},
  author  = {Bran, Andr{\'e}s M. and Cox, Sam and Schilter, Oliver and
             Baldassari, Carlo and White, Andrew D. and Schwaller, Philippe},
  journal = {Nature Machine Intelligence},
  volume  = {6},
  number  = {5},
  pages   = {525--535},
  year    = {2024},
  doi     = {10.1038/s42256-024-00832-8}
}

@article{lam2023graphcast,
  title   = {Learning Skillful Medium-Range Global Weather Forecasting},
  author  = {Lam, Remi and Sanchez-Gonzalez, Alvaro and Willson, Matthew and
             Wirnsberger, Peter and Fortunato, Meire and Alet, Ferran and
             Ravuri, Suman and Ewalds, Timo and Eaton-Rosen, Zach and Hu, Weihua and
             Merose, Alexander and Hoyer, Stephan and Holland, George and
             Vinyals, Oriol and Stott, Jacklynn and Pritzel, Alexander and
             Mohamed, Shakir and Battaglia, Peter},
  journal = {Science},
  volume  = {382},
  number  = {6677},
  pages   = {1416--1421},
  year    = {2023},
  doi     = {10.1126/science.adi2336}
}

@inproceedings{zelikman2022star,
  title     = {{STaR}: Bootstrapping Reasoning With Reasoning},
  author    = {Zelikman, Eric and Wu, Yuhuai and Mu, Jesse and Goodman, Noah},
  booktitle = {Advances in Neural Information Processing Systems},
  volume    = {35},
  pages     = {15476--15488},
  year      = {2022},
  publisher = {Curran Associates, Inc.},
  doi       = {10.52202/068431-1126}
}

@article{deepseekai2025r1,
  title   = {{DeepSeek-R1} Incentivizes Reasoning in {LLMs} through Reinforcement Learning},
  author  = {Guo, Daya and others},
  journal = {Nature},
  volume  = {645},
  number  = {8081},
  pages   = {633--638},
  year    = {2025},
  doi     = {10.1038/s41586-025-09422-z}
}

@article{yin2024survey,
  title={A Survey on Multimodal Large Language Models},
  author={Yin, Shukang and Fu, Chaoyou and Zhao, Sirui and Li, Ke and Sun, Xing and Xu, Tong and Chen, Enhong},
  journal={National Science Review},
  volume={11},
  number={12},
  pages={nwae403},
  year={2024},
  publisher={Oxford University Press}
}

@inproceedings{tan2025reasonrft,
  title     = {{Reason-RFT}: Reinforcement Fine-Tuning for Visual Reasoning of Vision Language Models},
  author    = {Tan, Huajie and Ji, Yuheng and Hao, Xiaoshuai and Chen, Xiansheng and Wang, Pengwei and Wang, Zhongyuan and Zhang, Shanghang},
  booktitle = {Advances in Neural Information Processing Systems},
  volume    = {38},
  pages     = {5772--5822},
  year      = {2025},
  publisher = {Curran Associates, Inc.},
  doi       = {10.52202/085713-0204}
}

@article{bai2025qwen3vl,
  author = {Bai, Shuai and Cai, Yuxuan and Chen, Ruizhe and others}, title = {{Qwen3-VL} Technical Report},
  journal = {arXiv preprint arXiv:2511.21631}, year = {2025}, url = {https://arxiv.org/abs/2511.21631}}

@article{zou2026interns1pro,
  author = {Zou, Yicheng and Zhu, Dongsheng and Zhu, Lin and others}, title = {{Intern-S1-Pro}: Scientific Multimodal Foundation Model at Trillion Scale},
  journal = {arXiv preprint arXiv:2603.25040}, year = {2026}, url = {https://arxiv.org/abs/2603.25040}}

@article{an2026llavaonevision2,
  author = {An, Xiang and Xie, Yin and Tang, Feilong and others}, title = {{LLaVA-OneVision-2}: Towards Next-Generation Perceptual Intelligence},
  journal = {arXiv preprint arXiv:2605.25979}, year = {2026}, url = {https://arxiv.org/abs/2605.25979}}

@misc{openai2026gpt55,
  author = {{OpenAI}}, title = {{GPT-5.5} Model},
  year = {2026}, howpublished = {OpenAI API documentation}, url = {https://developers.openai.com/api/docs/models/gpt-5.5}, note = {Accessed September 22, 2026}}

@misc{deepmind2025gemini3flash,
  author = {{Google DeepMind}}, title = {{Gemini 3 Flash} Model Card},
  year = {2025}, month = dec, url = {https://deepmind.google/models/model-cards/gemini-3-flash/}}

@misc{deepmind2026gemini31pro,
  author = {{Google DeepMind}}, title = {{Gemini 3.1 Pro} Model Card},
  year = {2026}, month = feb, url = {https://deepmind.google/models/model-cards/gemini-3-1-pro/}}

@misc{anthropic2026claudeopus47,
  author = {{Anthropic}}, title = {System Card: {Claude Opus 4.7}},
  year = {2026}, month = apr, url = {https://www-cdn.anthropic.com/037f06850df7fbe871e206dad004c3db5fd50340/Claude%20Opus%204.7%20System%20Card.pdf}}

@misc{qwen2026qwen36plus,
  author = {{Qwen Team}}, title = {{Qwen3.6-Plus}: Towards Real World Agents},
  year = {2026}, month = apr, howpublished = {Qwen Blog}, url = {https://qwen.ai/blog?id=qwen3.6}}

@article{bytedance2026seed20,
  author = {{ByteDance Seed}}, title = {{Seed2.0} Model Card: Towards Intelligence Frontier for Real-World Complexity},
  journal = {arXiv preprint arXiv:2607.00248}, year = {2026}, url = {https://arxiv.org/abs/2607.00248}}

@article{vteam2026glm5vturbo,
  author = {{V Team} and Hong, Wenyi and Gu, Xiaotao and others}, title = {{GLM-5V-Turbo}: Toward a Native Foundation Model for Multimodal Agents},
  journal = {arXiv preprint arXiv:2604.26752}, year = {2026}, url = {https://arxiv.org/abs/2604.26752}}

@misc{moonshot2026kimik26,
  author = {{Moonshot AI}}, title = {{Kimi K2.6}: Advancing Open-Source Coding},
  year = {2026}, howpublished = {Kimi Technical Blog}, url = {https://www.kimi.ai/blog/kimi-k2-6}, note = {Accessed September 22, 2026}}

@inproceedings{dosovitskiy2021vit,
  author = {Dosovitskiy, Alexey and Beyer, Lucas and Kolesnikov, Alexander and Weissenborn, Dirk and Zhai, Xiaohua and Unterthiner, Thomas and Dehghani, Mostafa and Minderer, Matthias and Heigold, Georg and Gelly, Sylvain and Uszkoreit, Jakob and Houlsby, Neil}, title = {An Image is Worth {16x16} Words: Transformers for Image Recognition at Scale},
  booktitle = {International Conference on Learning Representations}, year = {2021}, url = {https://arxiv.org/abs/2010.11929}}

@inproceedings{yue2025reasoningcapacity,
  title     = {Does Reinforcement Learning Really Incentivize Reasoning Capacity in LLMs Beyond the Base Model?},
  author    = {Yue, Yang and Chen, Zhiqi and Lu, Rui and Zhao, Andrew and Wang, Zhaokai and Yue, Yang and Song, Shiji and Huang, Gao},
  booktitle = {Advances in Neural Information Processing Systems},
  volume    = {38},
  pages     = {57654--57689},
  year      = {2025},
  publisher = {Curran Associates, Inc.},
  doi       = {10.52202/085713-1933}
}

@misc{deepmind2026gemini35flash,
  author = {{Google DeepMind}}, title = {{Gemini 3.5 Flash} Model Card},
  year = {2026}, month = may, url = {https://deepmind.google/models/model-cards/gemini-3-5-flash/}}

@article{wu2025propinn,
  title={Propinn: Demystifying propagation failures in physics-informed neural networks},
  author={Ma, Yuezhou and Wu, Haixu and Zhou, Hang and Weng, Huikun and Wang, Jianmin and Long, Mingsheng},
  journal={arXiv preprint arXiv:2502.00803},
  year={2025}
}

@InProceedings{pmlr-v267-luo25o,
  title = 	 {Transolver++: An Accurate Neural Solver for {PDE}s on Million-Scale Geometries},
  author =       {Luo, Huakun and Wu, Haixu and Zhou, Hang and Xing, Lanxiang and Di, Yichen and Wang, Jianmin and Long, Mingsheng},
  booktitle = 	 {Proceedings of the 42nd International Conference on Machine Learning},
  pages = 	 {41432--41449},
  year = 	 {2025}
}

@misc{chang2015shapenet,
      title={ShapeNet: An Information-Rich 3D Model Repository}, 
      author={Angel X. Chang and Thomas Funkhouser and Leonidas Guibas and Pat Hanrahan and Qixing Huang and Zimo Li and Silvio Savarese and Manolis Savva and Shuran Song and Hao Su and Jianxiong Xiao and Li Yi and Fisher Yu},
      year={2015},
      eprint={1512.03012},
      archivePrefix={arXiv}
}
\bibliographystyle{iclr2027_conference}
\newpage
\appendix
\appendix
\section{Full Benchmark Details}
\subsection{Task List and Benchmark Statistics}
\paragraph{Full task inventory.}
The benchmark contains 1,160 questions across 24 leaf tasks, with eight tasks in each field domain. Controlled equation fields contribute 370 questions, simulated physical fields contribute 410, and observed physical fields contribute 380. There are 4 mechanism-inference tasks, 12 control-inference tasks, and 8 outcome-inference tasks. Table~\ref{tab:appendix_task_inventory} lists every task, its target, inference type, post-training split, evaluation count, and number of choices.
The four mechanism tasks use a single image per question and account for 210 questions. The remaining twenty tasks use two images per question and account for 950 questions. Thus, the evaluation set contains 2,110 image references. An image may contain an entire spacetime heatmap, a montage of temporal snapshots, several CFD views, or multiple observational modalities.
\begin{table*}[h]
\centering
\caption{
Complete task inventory of PhysFieldBench.
M, C, and O denote mechanism, control, and outcome inference, respectively.
T and H denote training-covered and held-out tasks.
$n$ is the number of evaluation questions, $K$ is the number of answer choices,
and $r=1/K$ is the uniform random baseline.
Task IDs are used throughout the appendix.
}
\label{tab:appendix_task_inventory}
\small
\setlength{\tabcolsep}{2pt}
\renewcommand{\arraystretch}{0.95}
\begin{tabularx}{\textwidth}{@{}l X l cc rrc@{}}
\toprule
ID & Task name & Source & Axis & Split & $n$ & $K$ & $r$ \\
\midrule
\multicolumn{8}{@{}l}{\textit{Controlled equation fields}}\\
C01 &
Single Dominant Mechanism Identification &
PDEFormer1D &
M & T & 50 & 5 & $1/5$ \\
C02 &
Dominant Mechanism Pair Identification &
PDEFormer1D &
M & \emph{H} & 60 & 6 & $1/6$ \\
C03 &
Diffusion Coefficient Comparison &
PDEFormer1D &
C & T & 40 & 2 & $1/2$ \\
C04 &
Nonlinear Advection Coefficient Comparison &
PDEFormer1D &
C & T & 40 & 8 & $1/8$ \\
C05 &
Single Dominant Mechanism Identification &
PDEFormer2D &
M & T & 40 & 4 & $1/4$ \\
C06 &
Dominant Mechanism Pair Identification &
PDEFormer2D &
M & T & 60 & 6 & $1/6$ \\
C07 &
Diffusion Coefficient Comparison &
PDEFormer2D &
C & \emph{H} & 40 & 2 & $1/2$ \\
C08 &
Nonlinear Advection Coefficients Comparison &
PDEFormer2D &
C & T & 40 & 4 & $1/4$ \\
\midrule
\multicolumn{8}{@{}l}{\textit{Simulated physical fields}}\\
S01 &
Rayleigh \& Prandtl Number Comparison &
The Well &
C & T & 60 & 4 & $1/4$ \\
S02 &
Reynolds \& Schmidt Number Comparison &
The Well &
C & T & 60 & 4 & $1/4$ \\
S03 &
Radiative Cooling Time Comparison &
The Well &
C & T & 40 & 2 & $1/2$ \\
S04 &
Drag Coefficient Comparison &
ShapeNetCar &
O & T & 50 & 2 & $1/2$ \\
S05 &
Lift Coefficient Comparison &
ShapeNetCar &
O & T & 50 & 2 & $1/2$ \\
S06 &
Mach Number \& Angle-of-Attack Comparison &
NASA-CRM &
C & T & 50 & 4 & $1/4$ \\
S07 &
Drag Coefficient Comparison &
AirCraft &
O & \emph{H} & 50 & 2 & $1/2$ \\
S08 &
Lift Coefficient Comparison &
AirCraft &
O & \emph{H} & 50 & 2 & $1/2$ \\
\midrule
\multicolumn{8}{@{}l}{\textit{Observed physical fields}}\\
O01 &
Cylinder Reynolds Number Comparison &
RealPDEBench &
C & T & 40 & 2 & $1/2$ \\
O02 &
Controlled-Cylinder Oscillation Frequency Comparison &
RealPDEBench &
C & T & 40 & 2 & $1/2$ \\
O03 &
Controlled-Cylinder Reynolds Number Comparison &
RealPDEBench &
C & \emph{H} & 40 & 2 & $1/2$ \\
O04 &
Foil Reynolds Number \& Angle-of-Attack Comparison &
RealPDEBench &
C & T & 60 & 4 & $1/4$ \\
O05 &
Tropical Cyclone Size Comparison &
TCIR &
O & T & 50 & 2 & $1/2$ \\
O06 &
12-h Intensity Change Comparison &
TCIR &
O & T & 50 & 2 & $1/2$ \\
O07 &
12-h Intensity Change Comparison (Cross-Basin Set) &
TCIR &
O & \emph{H} & 50 & 2 & $1/2$ \\
O08 &
High-VIL Coverage Comparison &
SEVIR &
O & T & 50 & 2 & $1/2$ \\
\midrule
\multicolumn{5}{@{}l}{\textbf{Total}} &
\textbf{1,160} & & \\
\bottomrule
\end{tabularx}
\end{table*}
\paragraph{Dataset sources.}
The eight controlled-equation tasks use one- and two-dimensional scalar-field simulations organized through the PDEFormer~\citep{ye2024pdeformer} data construction pipeline. Both PDE families are solved using Dedalus v3 as a numerical solver to generate spatiotemporal field solutions. 
For the one-dimensional tasks, the candidate terms are expressed in the benchmark prompt as
\begin{equation}
 u_t+c_1u+c_2u^2+c_3u_x+c_4u u_x
 +c_5\partial_x\!\left(\kappa(x)u_x\right)=0, \quad (t,x)\in[0,1]\times[-1,1]
 \label{eq:appendix_pde_family}
\end{equation}
where $u=u(t,x)$ and subscripts denote partial derivatives. The terms $u$ and $u^2$ represent linear and quadratic reaction, $u_x$ and $u u_x$ represent linear and nonlinear advection, and $\partial_x(\kappa(x)u_x)$ is the diffusion operator. The coefficients $c_i$ are signed multipliers in this term-based notation.
For the two-dimensional tasks, the simplified generating equation is
\begin{equation}
\begin{aligned}
 &u_t+c_1u+c_{2x}u_x+c_{2y}u_y
 +u\left(c_{3x}u_x+c_{3y}u_y\right)
 =\kappa\left(u_{xx}+u_{yy}\right), \\
 &\quad (t,x,y)\in[0,1]\times[-1,1]\times[-1,1]
\end{aligned}
\label{eq:appendix_pde_2d}
\end{equation}
where $u=u(t,x,y)$. The coefficients $c_1$, $(c_{2x},c_{2y})$, and $(c_{3x},c_{3y})$ control reaction, directional linear advection, and directional nonlinear advection. The right-hand side represents isotropic diffusion with diffusivity $\kappa\geq0$, following the sign convention of the numerical solver. Directional terms are grouped into four mechanism classes: reaction, linear advection, nonlinear advection, and diffusion.
Labels are derived from the generating coefficients. Mechanism-recognition tasks identify one or two dominant terms or mechanism groups, while coefficient-comparison tasks ask about diffusion strength or nonlinear-advection coefficients. The one-dimensional nonlinear-advection task compares coefficient signs and magnitudes, and its two-dimensional counterpart compares the magnitudes in the two spatial directions.
The Well~\citep{ohana2024well} provides Rayleigh--B\'enard convection, shear flow, and turbulent radiative-layer simulations. Their targets are the Rayleigh and Prandtl numbers, Reynolds and Schmidt numbers, and cooling time, respectively. Engineering tasks use ShapeNetCar~\citep{chang2015shapenet}, NASA-CRM~\citep{bekemeyer2025introduction}, and AirCraft~\citep{pmlr-v267-luo25o} CFD data. ShapeNetCar and AirCraft require inference of integrated aerodynamic coefficients, while NASA-CRM requires joint comparison of Mach number and angle of attack. The ground truth is obtained from simulation parameters or stored aerodynamic coefficients.
RealPDEBench~\citep{hu2026realpdebench} provides experimental wake measurements for a stationary cylinder, a cylinder with imposed periodic motion, and a hydrofoil. Recorded experimental conditions supply the Reynolds number, imposed frequency, and angle-of-attack labels. TCIR~\citep{chen2018rotation} supplies tropical cyclone observations and associated attributes for size and future-intensity comparisons. SEVIR~\citep{veillette2020sevir} provides satellite observations as model inputs, while the radar VIL channel is used only to construct the ground truth labels. These tasks connect visible field structure to experimental settings, system attributes, or subsequently observed outcomes.
\paragraph{Task taxonomy.}
Tasks are classified along two complementary axes. The field domain identifies the origin of the observations: controlled equation fields are generated by parameterized PDEs, simulated physical fields come from fluid and engineering simulations, and observed physical fields come from experiments or natural observations. The inference axis identifies the requested target. Mechanism inference recovers the dominant generating terms or mechanisms; control inference recovers an ordering of parameters or operating conditions; outcome inference compares system responses or observed attributes, including aerodynamic coefficients and meteorological outcomes. Domain and inference type are assigned independently. The assignment of every leaf task is given in Table~\ref{tab:appendix_task_inventory}.
\subsection{Rendering and Visual Input Construction}
We render the physical observations before model evaluation and reuse the resulting images across models. The rendering follows the structure of the source data: spacetime heatmaps for one-dimensional dynamics, temporal montages for evolving two-dimensional fields, multi-view panels for engineering simulations, and multimodal panels for meteorological observations. Table~\ref{tab:appendix_visual_inputs} summarizes the input representations.
The one-dimensional heatmap exposes the full sampled evolution along one spatial coordinate. The two-dimensional montages retain spatial structure within frames and temporal order across frames. Engineering panels expose several views or field variables from one simulation case, while meteorological panels combine complementary modalities from the same scene. 
\begin{table}[t]
\centering
\caption{Visual inputs by task family. A paired question presents two complete images of the listed type in a fixed order.}
\label{tab:appendix_visual_inputs}
\small
\setlength{\tabcolsep}{5pt}
\renewcommand{\arraystretch}{1.15}
\begin{adjustbox}{max width=\linewidth}
\begin{tabular*}{\linewidth}{
  @{\extracolsep{\fill}}
  p{0.1\linewidth}
  p{0.33\linewidth}
  p{0.5\linewidth}
  @{}
}
\toprule
Tasks & Visible field & Image organization \\
\midrule
C01--C04 & Scalar field $u(t,x)$ & Spacetime heatmap with 256 spatial locations and 101 time steps. $x\in[-1,1]$ increases left to right; $t\in[0,1]$ increases top to bottom.\\
C05--C08 & Scalar field $u(t,x,y)$ & Twenty snapshots in a $4\times5$ chronological montage; $128\times128$ spatial grid and $t\in[0,1]$.\\
S01--S03 & Buoyancy; Passive tracer; Density & A $4\times5$ montage of twenty frames for each simulation trajectory.\\
S04--S05 & Car external-flow fields & Top view, center-$Y$ slice, and center-$Z$ slice of each car.\\
S06 & \raggedright Surface pressure coefficient and skin-friction magnitude & Multi-view pressure panels in the upper half and skin-friction panels in the lower half.\\
S07--S08 & \raggedright Surface pressure coefficient $C_p$ & Six views: front oblique, rear oblique, left, right, top, and bottom.\\
O01--O04 & Streamwise velocity $u$ & Twenty experimental wake snapshots in a $4\times5$ chronological montage; the main flow is left to right.\\
O05--O07 & IR1, WV, and PMW & Infrared, water-vapor, and passive-microwave panels of the current cyclone scene.\\
O08 & VIS, IR069, and IR107 & Three satellite channels at the current time.\\
\bottomrule
\end{tabular*}
\end{adjustbox}
\end{table}
\subsection{Question Templates and Label Spaces}
\paragraph{Question templates.}
Each task-specific prompt follows five components: task background, input-image description, physical or data background, the question with answer options, and the output instruction. The image description specifies the visible quantities, panel organization, image ordering, and relevant coordinates or time conventions. The physical background defines the quantities or candidate mechanisms needed to interpret the question, while the output instruction requests one uppercase option letter. Mechanism questions ask which generating term or mechanism is dominant, while comparison questions ask for the relative ordering of a specified control variable or outcome across the two cases.
\paragraph{Answer choices.}
Each question specifies a physical target and a finite answer space. Mechanism recognition is categorical. Continuous control variables and outcome quantities are converted into comparisons between two cases. Joint-comparison tasks combine several required judgments into one option, so a correct prediction must resolve every component.
For the one-dimensional single-term task C01, options A--E correspond to $u$, $u^2$, $u_x$, $uu_x$, and $(\kappa(x)u_x)_x$. C02 asks for the pair of dominant terms among $u$, $u_x$, $uu_x$, and $(\kappa(x)u_x)_x$. These represent reaction, linear advection, nonlinear advection, and diffusion. It is worth noting that $u^2$ is not an option in task C02. In two dimensions, directional transport terms are grouped into mechanisms: C05 selects one of reaction, linear advection, nonlinear advection, and diffusion, while C06 selects a pair from these four categories.
\begin{table}[htbp]
\centering
\caption{Option mapping for the two-mechanism tasks C02 and C06. In C02, the mechanism names refer to their one-dimensional terms.}
\label{tab:appendix_mechanism_options}
\small
\begin{adjustbox}{max width=\linewidth}
\begin{tabular*}{\linewidth}{@{\extracolsep{\fill}}cl@{}}
\toprule
Option & Dominant pair\\
\midrule
A & Reaction and linear advection\\
B & Reaction and nonlinear advection\\
C & Reaction and diffusion\\
D & Linear advection and nonlinear advection\\
E & Linear advection and diffusion\\
F & Nonlinear advection and diffusion\\
\bottomrule
\end{tabular*}
\end{adjustbox}
\end{table}
For a binary target $z$, option A indicates $z_1>z_2$ and option B indicates $z_1<z_2$, where subscripts identify the ordered input images. The target is a numerical value unless the question explicitly requests an absolute magnitude. 
For a joint target $(z,w)$, the four options are
\begin{equation}
\begin{aligned}
\mathrm{A}&:\ (z_1>z_2,\ w_1>w_2),&
\mathrm{B}&:\ (z_1>z_2,\ w_1<w_2),\\
\mathrm{C}&:\ (z_1<z_2,\ w_1>w_2),&
\mathrm{D}&:\ (z_1<z_2,\ w_1<w_2).
\end{aligned}
\label{eq:appendix_joint_options}
\end{equation}
The ordered targets are $(|c_x|,|c_y|)$ for C08, $(Ra,Pr)$ for S01, $(Re,Sc)$ for S02, $(Ma,\alpha)$ for S06, and $(Re,\alpha)$ for O04. C08 therefore compares directional coefficient magnitudes without asking for their signs. C04 additionally requires the sign of the nonlinear-advection coefficient in each one-dimensional field.
The Well labels compare its generating $(Ra,Pr)$ and $(Re,Sc)$ parameters or cooling time $t_{\rm cool}$. ShapeNetCar and AirCraft use CFD-derived drag and signed lift coefficients $C_d$ and $C_l$; NASA-CRM supplies the operating $(Ma,\alpha)$ pair. RealPDEBench labels come from recorded cylinder Reynolds numbers, imposed controlled-cylinder frequency, and the foil conditions $(Re,\alpha)$. TCIR supplies the size attribute for the cyclone-size task. Each comparison follows the binary or joint option convention above.
For the two future-intensity tasks, let $V_i(t)$ denote the stored cyclone-intensity attribute at the observation time $t$ in case $i$. The comparison target is the signed change
\begin{equation}
\Delta V_{12,i}=V_i(t+12\,\mathrm{h})-V_i(t).
\label{eq:appendix_intensification}
\end{equation}
The question asks which cyclone undergoes the larger intensity increase over the following twelve hours, and its ground truth selects the larger observed $\Delta V_{12}$. O06 and O07 use the same target with different basin subsets.
The SEVIR task compares the current spatial coverage of strong radar VIL echoes using only satellite inputs. If $\widetilde{\mathrm{VIL}}_i(p)$ is the encoded radar value at native-grid pixel $p$ in the selected current frame in case $i$, the target spatial coverage is
\begin{equation}
A_{133,i}=\sum_p \mathbf{1}\!\left[\widetilde{\mathrm{VIL}}_i(p)\geq133\right].
\label{eq:appendix_vil_area}
\end{equation}
\paragraph{Random baselines.}
For a task with $K$ valid options, uniform random choice has accuracy $r=1/K$. The benchmark includes fourteen binary tasks, six four-choice tasks, two six-choice tasks, one five-choice task, and one eight-choice task. Their random baselines are reported per task in Table~\ref{tab:appendix_task_inventory}. This baseline refers to uniform option sampling. We use the random baselines to account for differences in chance-level accuracy arising from different numbers of answer options.
\subsection{Pair Construction and Quality Control}
\paragraph{Pairwise target margins.}
For comparison tasks, candidate pairs are formed using the physical values in the construction metadata. A pair is retained only when the target ordering is well-defined and its gap satisfies the task-specific selection rule. Where the metadata permits, other conditions are matched to reduce variation unrelated to the question. Table~\ref{tab:appendix_pair_rules} summarizes the selection rules.
\begin{table}[t]
\centering
\caption{Summary of selection rules. Gaps use source metadata units and angular gaps are in degrees. All cases are manually reviewed for interpretable visual cues, label consistency, and answer leakage.}
\label{tab:appendix_pair_rules}
\small
\setlength{\tabcolsep}{5pt}
\renewcommand{\arraystretch}{1.12}
\begin{adjustbox}{max width=\linewidth}
\begin{tabular*}{\linewidth}{@{\extracolsep{\fill}}p{0.2\linewidth}p{0.74\linewidth}@{}}
\toprule
Tasks & Target separation and matched conditions\\
\midrule
C01 / C02 / C05 / C06 & One or two dominant terms or mechanisms, with coefficient magnitudes at least $10\times$ those of non-dominant terms or mechanisms. Coefficient separation is used only for candidate screening. Final inclusion requires manual verification that the rendered evolution provides discernible physical evidence consistent with the intended mechanism label. Coefficient ratios alone do not establish dynamical dominance.\\
C03 / C07 & Both diffusion coefficients are positive, and $|\kappa_1-\kappa_2|\in[0.0285,0.0315]$ for 1D and $[0.00275,0.00325]$ for 2D.\\
C04 & Nonlinear-advection coefficient magnitudes differ by approximately $0.74$ or more.\\
C08 & Absolute coefficient magnitudes differ by $[0.01,0.05]$ in both spatial directions.\\
S01 & Distinct $Ra$ and $Pr$. Construction additionally excludes equal products and equal ratios.\\
S02 & Distinct $Re$ and $Sc$. Construction additionally excludes equal products.\\
S03 & Cooling times selected from $\{0.03,0.06,0.10,0.18,0.32,0.56,1.00,1.78,3.16\}$, and exclude adjacent levels.\\
S04 & $|C_{d,1}-C_{d,2}|\in[0.04,0.08]$.\\
S05 & Same sign $C_l$ and $|C_{l,1}-C_{l,2}|\in[0.12,0.20]$.\\
S06 & $|Ma_1-Ma_2|\geq0.20$ and $|\alpha_1-\alpha_2|\geq2.0^\circ$.\\
S07 & $|C_{d,1}-C_{d,2}|\in[0.02,0.04]$; fixed $Ma=7$, $\alpha=0^\circ$, and $\beta=0^\circ$.\\
S08 & $|C_{l,1}-C_{l,2}|\in[0.12,0.20]$; fixed $Ma=7$, $\alpha=7^\circ$, and $\beta=0^\circ$.\\
O01 / O03 & Larger-to-smaller Reynolds ratio in $(1.5,2.0)$. O03 also matches imposed frequency.\\
O02 & Equal Reynolds number and frequency difference in $(0.11,0.39)$.\\
O04 & Larger-to-smaller Reynolds ratio larger than $3$ and angle gap in $(0.1^\circ,14.9^\circ)$.\\
O05 & Same basin, size gap at least 40, and current-intensity gap at most 5.\\
O06 / O07 & $|\Delta V_{12,1}-\Delta V_{12,2}|\in[12,20]$, and same basin, different storms, current-intensity gap at most 5, and current-size gap at most 40.\\
O08 & Daytime scenes and $|A_{133,1}-A_{133,2}|\in[1500,3000]$ native-grid pixels.\\
\bottomrule
\end{tabular*}
\end{adjustbox}
\end{table}
\paragraph{Nuisance-factor constraints.}
The matching rule follows the target of each task. For example, the controlled-cylinder frequency task fixes Reynolds number, whereas the controlled-cylinder Reynolds task fixes frequency. The AirCraft tasks fix the operating condition so that the comparison concerns the aerodynamic response of different geometries. TCIR matching controls basin and current attributes before comparing size or future change. These constraints reduce specified confounds, while the remaining coupled physical effects are part of the inference challenge.
\paragraph{Manual inspection.}
The manual review was conducted by researchers with relevant expertise in physical-field analysis. We combine manual review with metadata checks and image-duplicate detection. Reviewers inspected all evaluation examples for corrupted or degenerate renderings, visually indistinguishable pairs, inconsistent labels, and residual answer-revealing cues. They also assessed whether the rendered fields contained discernible physical cues relevant to the queried target, flagging cases where the visual evidence appeared insufficient or ambiguous. Image hashes aided duplicate detection, while physical metadata established the ground-truth ordering. We applied target-separation thresholds alongside visual inspection because a numerical gap alone does not ensure that the rendered fields provide informative evidence.
\paragraph{Answer balancing.}
Answer distributions are balanced within each task to reduce answer-position bias, subject to data availability, with residual imbalance in S06. The random baseline is uniform selection over the valid options, $1/K$.
\subsection{Training-covered and Held-out Task Split}
\label{app:task_split}
\paragraph{Split definition.}MLLM post-training uses eighteen training-covered tasks: C01, C03--C06, C08, S01--S06, O01, O02, O04--O06, and O08. The six held-out tasks are C02, C07, S07, S08, O03, and O07. The corresponding evaluation subsets contain 870 and 290 questions, respectively. The independently supervised per-task ViT follows a different training protocol and is not subject to this MLLM task split.
\paragraph{Held-out selection rationale.}
The split is defined at the leaf-task level rather than by withholding an entire source dataset. Two tasks are left out in each field domain, with one mechanism-inference task, two control-inference tasks, and three outcome-inference tasks overall. This design retains related physical systems or inference operations among the covered tasks while withholding selected combinations of target, representation, or source setting. Related training and evaluation tasks may therefore share source data and physical abilities.
\paragraph{Atomic ability transfer.}
The split tests several forms of transfer. C02 combines related mechanism-recognition abilities in a held-out one-dimensional two-term task. C07 tests diffusion comparison in two-dimensional fields. S07 and S08 transfer aerodynamic inference to aerodynamic coefficients on AirCraft. O03 changes the queried control variable within experimental controlled-cylinder wakes. O07 transfers future-intensity comparison from ATL/WPAC to IO/SH cyclones. These settings examine whether learned mechanism recognition, visual comparison, and parameter interpretation transfer to a different task formulation or physical system.
\newpage
\section{Evaluation Protocol}
\subsection{Unified VQA Evaluation}
\paragraph{Evaluation format.}
All 24 tasks use a unified single-choice visual question answering format.
Each example consists of one or two rendered case images and a task-specific
question with a predefined set of answer options. The prompt describes the
physical background, the visualized quantities and image layout, and the
inference target. The model processes the complete example and selects one
option. In particular, a pairwise comparison is evaluated as one question
about two cases. Ground-truth labels and construction metadata are used for
scoring and are not included in the model input. The main evaluation uses
zero-shot prompting, while few-shot prompting is examined separately.
\vspace{-2pt}
\paragraph{Input packaging.}
All MLLMs receive the ordered case images together with the corresponding
question in a single multimodal request. The names \texttt{image1} and
\texttt{image2} in the prompt refer to the first and second image, respectively.
The visual inputs have the following organizations. Single-image and pairwise
formats specify the number of case images, whereas temporal grids, multiple
views, and multiple modalities describe the content within each image.
\noindent\textit{Single-image inputs.}
The four mechanism-recognition tasks provide one image of a physical
realization. For one-dimensional PDEs, this image is a spacetime heatmap,
with space along the horizontal axis and time increasing downward. For
two-dimensional PDEs, it is a montage of field snapshots at successive times.
The model identifies the dominant mechanism or mechanism combination from
this single case image.
\noindent\textit{Pairwise-image inputs.}
The remaining tasks provide two separate images in a fixed order. Each image
represents one case, and both are presented in the same request so that the
model can compare the relevant visual evidence. The question asks for an
ordering of a physical parameter or outcome, or a joint ordering of two
quantities. Each case image retains its internal panel organization. For
example, comparing two temporal sequences means supplying two montages.
\noindent\textit{Temporal-grid inputs.}
For two-dimensional PDEs and The Well, 20 snapshots from one trajectory are
arranged in a $4\times5$ chronological grid. Experimental-wake sequences are
also rendered as temporal montages. Each complete grid is supplied as one
image, allowing the model to assess evolution across panels. Thus, a paired
sequence comparison uses two image attachments, each containing the frames
of its own trajectory. The prompt explains the temporal ordering and the
visualized field. These inputs are processed through the models' image
interfaces. One-dimensional spacetime heatmaps instead encode temporal
evolution directly along an image axis.
\noindent\textit{Multi-view inputs.}
Engineering tasks combine complementary views of the same case into a single
composite image. ShapeNetCar includes a top view and central slices. AirCraft
includes six views of the surface pressure coefficient. NASA-CRM combines multiple
surface-pressure and skin-friction views. A paired question supplies two
such composites. View definitions and visualized quantities are described
in the prompt, enabling comparison of local structures across viewpoints
and integration of evidence at the case level.
\noindent\textit{Multimodal inputs.}
Meteorological tasks combine three observation modalities into one image
per case. TCIR includes infrared, water-vapor, and passive-microwave panels,
while SEVIR includes VIS, IR069, and IR107 panels. A pairwise question therefore
provides two composite images, each containing the three modalities of its
respective scene. The prompt identifies the channels and the target quantity.
The case-level organization is shared across MLLMs, while each model's
processor or API handles visual encoding and its native multimodal prompt
format. Local Qwen checkpoints use their associated processor and chat
template. API models receive the question and ordered image attachments
through their multimodal interface. The supervised ViT
baseline uses a task-specific visual classification interface, detailed in
Appendix~\ref{app:vit_baseline}.
\vspace{-2pt}
\paragraph{Output requirement.}
The standard task prompt requests exactly one uppercase letter from the valid
option set, with no additional text. Tasks may have two, four, five, six, or eight
options. For models that produce reasoning, evaluation extracts the answer
option from the response and compares it with the ground-truth label.
All accuracy metrics use this option-level decision, and explanation quality is assessed separately in the self-explanation analysis.
\subsection{Prompting and Answer Parsing}
\paragraph{Standard prompt.}
The main evaluation uses task-specific zero-shot prompts with a common structure:
task background, input-image description, physical or data background, question
and answer options, and output format. All prompts contain the following instruction:
\begin{quote}
\small\ttfamily\raggedright
Use only the visual content and the information explicitly provided in this prompt.
\end{quote}
For a binary task, the output instruction is:
\begin{quote}
\small\ttfamily\raggedright
Your response must be exactly one letter from: A, B
Do not output anything else.
\end{quote}
The permitted letters are adjusted to the task's answer space. Representative zero-shot task prompts, including their physical context, image descriptions, answer options, and output instructions, are provided in Appendix~\ref{app:exact_prompts}.
\paragraph{Answer extraction.}
We use deterministic, rule-based parsing to map each response to a valid answer
option. For the local base and post-trained models, delimited thinking content
is removed before parsing. The parser first accepts a response consisting of
a valid letter, then checks an option at the beginning or end of a response line,
and finally applies letter-matching fallbacks. Extracted letters are normalized
to uppercase and checked against the task's option set. The API evaluator
prioritizes explicit answer tags before applying answer-line and option-matching
rules. The resulting option is compared with the stored ground truth.
\paragraph{Invalid responses.}
Responses from which no valid option can be extracted are counted as incorrect
and remain in the evaluation denominator. Multiple conflicting answers are treated as invalid. Responses truncated at the generation limit are also considered invalid and counted as incorrect if no answer option can be extracted from the beginning. Generation budgets and truncation counts are reported in
Appendix~\ref{app:inference_settings}.
\subsection{Metrics}
For task $t$ with $n_t$ cases and $K_t$ options, let $c_t$ be the number of correct
parsed answers, counting invalid and strict-truncated responses as incorrect.
Task accuracy and normalized score are
\begin{equation}
 a_t=\frac{c_t}{n_t},\qquad
 s_t=100\frac{a_t-1/K_t}{1-1/K_t}.
 \label{eq:app_score}
\end{equation}
For a set of tasks $G$, we report
\begin{equation}
 S_G=\frac{1}{|G|}\sum_{t\in G}s_t
\end{equation}
All main normalized aggregates use equal task weights, including the domain,
inference-axis, covered, and held-out columns. A score of zero indicates performance equivalent to uniform random choice, while negative scores indicate performance worse than random choice. A score of 100 represents perfect accuracy. The minimum task score is $-100/(K_t-1)$, so a normalized score should not be interpreted as the percentage of correctly answered questions.
\subsection{Model Inference Settings}
\label{app:inference_settings}
\paragraph{Local models.}
The locally evaluated baselines are Qwen3-VL-8B-Instruct and LLaVA-OneVision-2-8B-Instruct; our post-training experiments start from Qwen3-VL-8B-Thinking. For the main single-response evaluation, all local models use greedy decoding. The two instruct baselines are evaluated in non-thinking mode with a 64-token generation limit. The Qwen3-VL-8B-Thinking base checkpoint is evaluated with thinking enabled to preserve its native reasoning behavior. CoT-SFT and GRPO are likewise evaluated with thinking enabled and a 9,216-token generation limit. In contrast, answer-only SFT trains the model to produce a final option directly. The resulting checkpoint emits a single option letter even when thinking is enabled; evaluating it with thinking disabled therefore does not remove a reasoning trace it would otherwise generate. Its 64-token limit is sufficient, with no truncated responses observed. All post-trained models use a maximum of 589,824 pixels per image, with resizing handled by the model processor. Table~\ref{tab:app_decoding} summarizes these settings.
\begin{table}[t]
\centering\small
\caption{Inference settings for the selected post-trained Qwen3-VL-8B-Thinking
models. The token limit includes all generated reasoning and answer text.
Image resolution is specified as a pixel budget rather than a fixed shape.}
\label{tab:app_decoding}
\begin{tabular*}{\textwidth}{@{\extracolsep{\fill}}lccc}
\toprule
Setting & Answer-only SFT & CoT SFT & CoT + GRPO \\
\midrule
Checkpoint & Final, epoch 5 & Step 185 & Step 320 \\
Decoding & Greedy & Greedy & Greedy \\
Thinking & Disabled & Enabled & Enabled \\
Maximum image pixels & 589,824 & 589,824 & 589,824 \\
Maximum generated tokens & 64 & 9,216 & 9,216 \\
Truncated responses & 0 & 8 & 4 \\
\bottomrule
\end{tabular*}
\end{table}
\paragraph{API-evaluated models.}
The API-based evaluation includes Gemini 3 Flash, Gemini 3.5 Flash, Gemini 3.1 Pro, GPT-5.5, Claude Opus 4.7, Qwen3.6 Plus, Doubao Seed 2.0 Pro, GLM-5V-Turbo, Kimi K2.6, and Intern-S1-Pro. Kimi K2.6 and Intern-S1-Pro are included in this group because the reported evaluations were conducted through APIs rather than local deployments. Images and questions are submitted through the respective multimodal interfaces. For models that expose a configurable reasoning-effort setting, the reasoning level is set to \texttt{high}. All API-based models use a maximum output length of 16,384 tokens and a temperature of 0 to provide sufficient reasoning capacity while improving inference reproducibility.
\section{Post-training Details}
In the main text, \emph{Answer Supervision} refers to supervised
fine-tuning (SFT) on the ground-truth final option alone, without a
reasoning trace; we call this \emph{answer-only SFT} below.
\emph{Reasoning Supervision} refers to SFT on a structured
chain-of-thought (CoT) rationale followed by the final option; we
call this \emph{CoT-SFT}. \emph{Reinforcement Learning} refers to
GRPO applied after CoT-SFT. Answer-only SFT and CoT-SFT start
independently from the same Qwen3-VL-8B-Thinking base model, while
GRPO continues training from the CoT-SFT model.
\subsection{Answer-only SFT}
\paragraph{Training data.}
We construct answer-only SFT data from the eighteen training-covered tasks
defined in Table~\ref{tab:appendix_task_inventory}. The examples draw on the same underlying
physical data collections as PhysFieldBench: parameterized PDE simulations,
The Well, ShapeNetCar and NASA-CRM simulations, RealPDEBench experiments, and
TCIR and SEVIR observations. Training examples are assembled into a separate
case collection using the corresponding task definitions and visual
representations. Mechanism-recognition examples associate a rendered field
with its generating mechanism label. Comparison examples pair two cases and
derive the labels from their recorded parameters or target quantities.
The SFT training set contains 4,500 examples, with 250 examples sampled from each of the eighteen covered tasks. The six held-out tasks contribute no training examples, allowing us to separately evaluate performance on covered tasks and transfer to unseen task formulations. For covered tasks, the training and test sets target the same physical inference problems but are constructed from separate case collections. Sampling is stratified by answer option within each task to maintain approximately balanced option frequencies. Each example consists of the task images and question as the user input, followed by the correct uppercase option as the assistant response. Training minimizes autoregressive cross-entropy on the answer response. 
\paragraph{Training setup.}
Training starts from Qwen3-VL-8B-Thinking and updates the language-model parameters
with full-parameter SFT. The vision tower and multimodal projector are frozen.
The configuration uses learning rate $10^{-5}$, a cosine schedule and warmup ratio
0.1. The effective batch size is 32.
The maximum image size is 589,824 pixels, sequence cutoff is 2,048 tokens, and
thinking is disabled. Training runs for 5 epochs.
\subsection{CoT SFT}
\paragraph{Teacher reasoning generation.}
The teacher pipeline uses Qwen3.6 Plus to generate task-specific reasoning, prioritizing traces that independently reach the correct answer. Because the teacher's low accuracy limits the collection of correct traces, we supplement these with answer-conditioned generation, providing the correct option to the teacher. We use  Gemini 3 Flash to standardize retained traces into structured explanations. Finally the clean CoT set contains 1,012 records.
\paragraph{Structured reasoning template.}
The intended six components are visual observations, physical mapping,
evidence integration, pairwise comparison, factor disentanglement, and target
alignment. The rewrite pipeline also permits an alternative-answer check, as
shown in the self-explanation prompt in Appendix~\ref{app:self_explain_prompt}.
These components aim to distinguish a statement about visible structure from
the physical rule used to interpret it. The target text is serialized with
\texttt{<think>... </think>} reasoning and a final option.
\paragraph{Training configuration.}
The CoT-SFT model also uses a learning rate of $10^{-5}$ with a cosine
schedule and a warmup ratio of 0.1. The effective batch size is 32. The maximum image resolution is limited
to 589,824 pixels and the sequence cutoff is 4,096 tokens. Note that the
thinking template is enabled. As in answer-only
SFT, the language model is trained while the vision tower and projector are
frozen. We use the checkpoint at training step 185.
\paragraph{Teacher-data limitations.}
Filtering for correct answers may favor examples that the teacher
finds easier, and correct final answers do not guarantee physically
valid intermediate reasoning. The resulting rationales therefore
provide imperfect supervision.
\subsection{Reinforcement Learning}
\paragraph{RL data construction.}
During the RL stage, we use 2,000 examples selected
from the 4,500-example pool of eighteen training-covered tasks.
After excluding those examples used for CoT-SFT,
we select from the remaining candidates with nearly equal
task quotas and balanced answer options within each task.
Each record provides the images, prompt and answer for reward computation. No held-out task
contributes training examples.
\paragraph{Reward design.}
For a valid parsed option $\hat y$ in a $K$-choice task, the chance-normalized
reward is
\begin{equation}
 r_K(\hat y,y)=
 \begin{cases}
 1,&\hat y=y,\\
 -1/(K-1),&\hat y\in\mathcal{Y}_t,\ \hat y\ne y,\\
 -1,&\text{invalid output}.
 \end{cases}
 \label{eq:app_reward}
\end{equation}
Its expectation under uniform valid-option sampling is zero, and for valid
answers it is the per-example analogue of Equation~\ref{eq:app_score} divided by
100. Invalid outputs receive a stronger penalty than valid wrong answers when
$K>2$.
\paragraph{GRPO configuration.}
We report rollout number, batch size, learning rate, KL coefficient, maximum response length, and validation frequency.
The policy is initialized from the CoT-SFT checkpoint and trained
with a batch size of 64, a PPO minibatch size of 64, eight rollouts
per prompt, a learning rate of $5\times10^{-6}$, and a KL
regularization coefficient of 0.01. The maximum total sequence length is 4,096 tokens. We report results
from the checkpoint at step 320.
The group-relative update increases the relative probability of responses that
obtain higher answer rewards within a prompt group, subject to the configured
KL regularization.
\paragraph{Interpretation of stagewise comparisons. }Our post-training analysis focuses on changes within training paths, rather than an equal-budget ranking of supervision objectives. Relative to the base checkpoint, answer-only SFT substantially improves performance on training-covered tasks (-8.1 to 58.2), showing that direct-answer training enables the model to solve many tasks represented in its training data. Starting from the CoT-SFT checkpoint, GRPO further improves held-out-task performance (22.3 to 33.3) under the same thinking-enabled evaluation setting; all six held-out tasks improve (Tables 14--16). This result characterizes the effect of the GRPO training stage conditional on CoT-SFT initialization. Because the paths differ in data construction, training volume, and response format, we do not interpret score differences between the answer-only SFT and CoT-SFT-GRPO checkpoints as an isolated causal effect of one supervision signal over another.
\section{Self-explanation Error Analysis}
\label{appendix:self_explain}
\subsection{Self-explanation Setup}
\paragraph{Model and sample selection.}
We use Gemini 3 Flash to generate self-explanations and analyze cases whose
parsed answer differs from the ground-truth option. The diagnostic sample
contains 162 incorrectly answered examples spanning all 24 tasks. The final
sample includes a saved response and a completed judgment for every example.
Sampling across tasks provides coverage of different physical systems and
inference targets, while restricting the analysis to failures allows us to
examine where the explanations break down. The resulting statistics characterize the selected failure cases and
provide insight into error patterns across PhysFieldBench. GPT-5.5 assesses the explanations using the category and rubric
in Appendix~\ref{app:judge}.
\paragraph{Self-explanation prompt.}
\label{app:self_explain_prompt}
We retain the original images, physical background, question, and answer
options, and replace the standard output instruction with the structured
prompt below. The prompt separates visible evidence from physical
interpretation and asks the model to integrate cues across the relevant
images, views, channels, or time steps. It also requests an alternative check
and a final answer, enabling the explanation and the selected option to be
examined together. Figure~\ref{fig:explain_prompt} shows the implemented output instruction.
\begin{figure}[htbp]
\centering
\includegraphics[width=\linewidth]{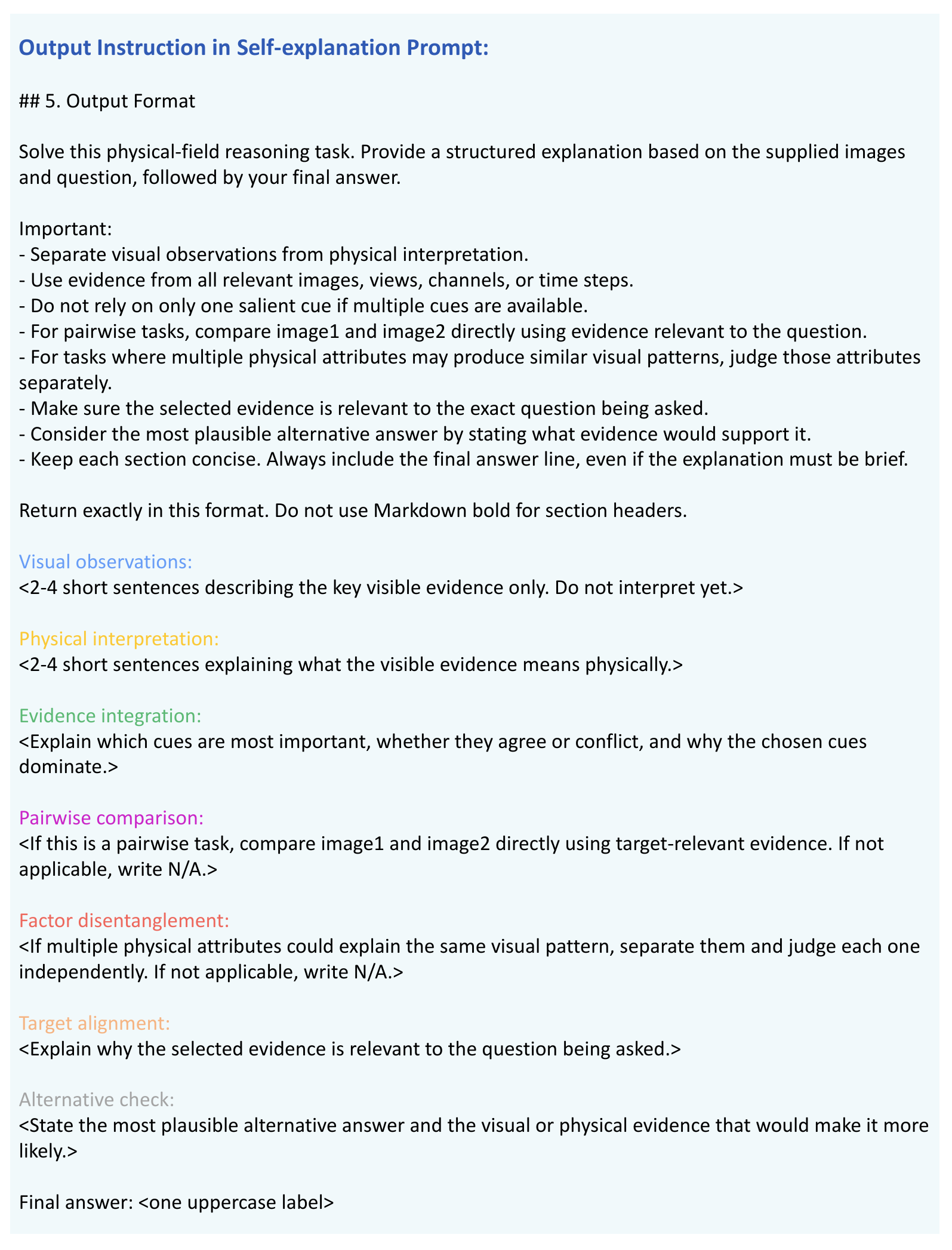}
\caption{Structured self-explanation prompt used in the diagnostic
analysis. It elicits six reasoning components, an alternative check,
and a final answer while retaining the original task context and
visual inputs.}
\label{fig:explain_prompt}
\end{figure}
\paragraph{Reasoning stages.}
We decompose each explanation into six interpretable components aligned with
the structured prompt:
\begin{enumerate}
\item \textbf{Visual observation:} identifying the relevant visible structures
and changes, such as spatial patterns, temporal evolution, or differences
between channels, before assigning them a physical meaning.
\item \textbf{Physical mapping:} connecting the observed cues to a
physical mechanism, parameter, or target quantity. This stage distinguishes
correct perception from a correct physical interpretation of that perception.
\item \textbf{Evidence integration:} combining multiple relevant cues,
assessing whether they agree or conflict, and explaining which evidence
supports the conclusion.
\item \textbf{Pairwise comparison:} directly comparing the two cases using
evidence relevant to the queried quantity and deriving the corresponding
relative ordering.
\item \textbf{Factor disentanglement:} distinguishing physical attributes
that can produce similar visual patterns and assessing the requested
attributes separately in tasks involving multiple factors.
\item \textbf{Target alignment:} checking that the selected evidence and
conclusion address the requested physical
quantity.
\end{enumerate}
Pairwise comparison and factor disentanglement may be marked not applicable
when the task does not require them. The alternative check supplements the
explanation but is not a seventh scored component. This decomposition provides a consistent framework for localizing
failures along the reasoning process, distinguishing errors in visual
perception from those in physical mapping, evidence integration,
and task-specific inference. It also supports comparisons of failure
patterns across tasks and physical domains, helping identify shared
bottlenecks in physical-field understanding.
\subsection{GPT-based Judge}
\label{app:judge}
\paragraph{Judge model.}
We use GPT-5.5 as a multimodal judge for the 162 failure cases.
Given the task images, question, options, model response, and
ground-truth answer with available target metadata, it assigns
error categories and scores the six reasoning components.
\paragraph{Error taxonomy.}
We define six error categories corresponding to the six reasoning
components above, allowing failures to be attributed to specific
aspects of physical-field reasoning:
\begin{enumerate}
\item \textbf{Visual pattern error (E1):} misreading or omission of decisive
visible structure, a channel, a view, motion, or temporal change.
\item \textbf{Physical mapping error (E2):} a noticed cue is mapped to the
wrong mechanism, coefficient, parameter, or physical meaning.
\item \textbf{Evidence integration error (E3):} anchoring on one salient
cue while failing to combine relevant evidence across fields, views,
channels, or time.
\item \textbf{Pairwise comparison error (E4):} a relative order fails at the comparison stage, rather than merely inheriting
an upstream mapping error.
\item \textbf{Factor disentanglement error (E5):} failure to distinguish
coupled attributes that can produce similar visible patterns.
\item \textbf{Target alignment error (E6):} addressing a related but
different quantity, such as present organization instead of future
intensity change.
\end{enumerate}
The judge selects one to three categories that materially contribute to the
failure and ranks them by contribution. The highest-ranked category
defines the primary error, while all selected categories contribute to the
multi-label statistics. The five-category visualization in Figure~\ref{fig:error_analysis}
combines pairwise-comparison and target-alignment errors into ``Other'', while the appendix
retains all six categories.
\paragraph{Scoring rubric.}
The rubric assigns a score of 0, 1, or 2 to each of the six reasoning components defined above, with higher scores indicating better satisfaction of the corresponding
criterion. For example, visual observation assesses the correctness and sufficiency of
visible evidence and physical mapping assesses its interpretation in relation to
the ground-truth target. Each component is scored against its corresponding criterion, with
a brief justification and supporting evidence. Components appropriately marked as not applicable receive full credit.
\subsection{Full Error Statistics}
\paragraph{Overall distribution.}
Table~\ref{tab:app_errors_overall} reports primary and multi-label error
frequencies over the 162 judged failures. Primary percentages use one category
per example and sum to 100\% up to rounding. Multi-label percentages measure
the fraction of examples assigned each category and can sum to more than 100\%.
Physical mapping is the most frequent primary error (53.09\%), followed by
visual pattern errors (36.42\%). Evidence integration is a primary error in
only 3.09\% of cases but contributes to 37.65\% under multi-label attribution,
showing that it often co-occurs with other failures.
Table~\ref{tab:app_reasoning_scores} reports the scores for all six reasoning
components, with a mean total of 5.90. Evidence integration receives the
lowest mean score, followed by pairwise comparison and
physical mapping. These results demonstrate that combining visual cues and translating them into physical judgments remain
weak points in the sampled explanations. Target alignment is comparatively
high, suggesting that the model generally
addresses the requested physical quantity, even when its reasoning
and final answer are incorrect.
\begin{table}[t]
\centering\small
\caption{Error attribution for 162 selected Gemini-3-Flash failures, judged by GPT-5.5. Primary counts include only the highest-ranked error per case, and multi-label counts include all assigned errors.}
\label{tab:app_errors_overall}
\begin{tabular*}{\linewidth}{@{\extracolsep{\fill}}lrrrr@{}}
\toprule
Type & Primary & Primary (\%) & Multi-label & Multi-label (\%) \\
\midrule
E1: Visual observation & 59 & 36.42 & 88 & 54.32 \\
E2: Physical mapping & 86 & 53.09 & 121 & 74.69 \\
E3: Evidence integration & 5 & 3.09 & 61 & 37.65 \\
E4: Pairwise comparison & 1 & 0.62 & 8 & 4.94 \\
E5: Factor disentanglement & 10 & 6.17 & 42 & 25.93 \\
E6: Target alignment & 1 & 0.62 & 2 & 1.23 \\
\bottomrule
\end{tabular*}
\end{table}
\begin{table}[tbp]
\centering
\small
\caption{Scores for 162 selected Gemini-3-Flash failures. Entries are means over individual cases. Higher scores indicate better satisfaction of each criterion. Each component is rated 0--2, and appropriately non-applicable comparison and disentanglement components receive 2 points.}
\label{tab:app_reasoning_scores}
\begin{tabular*}{\linewidth}{@{\extracolsep{\fill}}lr@{}}
\toprule
Reasoning component & Mean score \\
\midrule
Visual observation & 0.90 \\
Physical mapping & 0.68 \\
Evidence integration & 0.35 \\
Pairwise comparison & 0.67 \\
Factor disentanglement & 1.39 \\
Target alignment & 1.91 \\
\midrule
Total & 5.90 \\
\bottomrule
\end{tabular*}
\end{table}
\paragraph{Domain-level breakdown.}
Table~\ref{tab:app_error_groups} separates the cohort into 57 controlled-equation,
54 simulated-field, and 51 observed-field failures. Physical mapping is the leading primary error in all three domains. Beyond this shared difficulty, evidence
integration is implicated in 56.9\% of observed-field failures. Simulated
fields show the largest share of disentanglement errors (40.7\%), consistent
with the coupled-parameter engineering and fluid tasks. These patterns suggest that the main difficulties in physical-field understanding and reasoning vary across domains.
\paragraph{Inference-level breakdown.}
Mechanism, control, and outcome inference contribute 30, 89, and 43 failures.
Physical mapping errors occur in 29 of 30 mechanism-inference failures,
indicating a recurring difficulty linking field patterns to their generating
dynamics. Control inference contains all ten primary disentanglement errors,
consistent with tasks that require separating multiple physical parameters.
For outcome inference, integration errors occur in 26 of 43 failures (60.5\%),
highlighting difficulty combining cues to infer the requested outcome.
Thus, beyond the shared difficulty in physical mapping, control inference
poses challenges in parameter separation, while outcome inference
poses challenges in evidence integration.
\begin{table}[t]
\centering
\caption{Error counts by domain and inference axis in the selected failure cohort. Each E1--E6 cell is primary/multi-label count. Multi is the percentage of examples assigned more than one error.}
\label{tab:app_error_groups}
\small
\setlength{\tabcolsep}{3pt}
\begin{adjustbox}{max width=\linewidth}
\begin{tabular*}{\linewidth}{@{\extracolsep{\fill}}lrccccccr@{}}
\toprule
Group & $n$ & E1 & E2 & E3 & E4 & E5 & E6 & Multi (\%) \\
\midrule
Controlled & 57 & 25/37 & 27/41 & 0/13 & 1/4 & 4/13 & 0/0 & 84.2 \\
Simulated & 54 & 11/23 & 32/45 & 5/19 & 0/2 & 6/22 & 0/0 & 92.6 \\
Observed & 51 & 23/28 & 27/35 & 0/29 & 0/2 & 0/7 & 1/2 & 92.2 \\
\midrule
Mechanism & 30 & 12/19 & 18/29 & 0/5 & 0/0 & 0/5 & 0/0 & 93.3 \\
Control & 89 & 31/42 & 47/59 & 0/30 & 1/6 & 10/37 & 0/0 & 86.5 \\
Outcome & 43 & 16/27 & 21/33 & 5/26 & 0/2 & 0/0 & 1/2 & 93.0 \\
\midrule
All & 162 & 59/88 & 86/121 & 5/61 & 1/8 & 10/42 & 1/2 & 89.5 \\
\bottomrule
\end{tabular*}
\end{adjustbox}
\end{table}
\paragraph{Task-level breakdown.}
Table~\ref{tab:app_error_tasks} reports the number of judged failures, primary
and multi-label category counts, and mean rubric score for each task. Physical mapping
errors occur in every sampled two-term mechanism failure in both 1D and 2D
(C02 and C06). Mapping and disentanglement co-occur in all eight sampled cases
for both Rayleigh--B\'enard convection (S01) and NASA-CRM (S06), identifying
coupled-parameter interpretation as a recurring difficulty. In contrast,
visual pattern errors are primary in all five controlled-cylinder frequency
cases (O02), while mapping errors occur in all  12-hour cyclone
intensity-change failures (O06 and O07). These examples distinguish
perceptual difficulties from failures to infer physical properties or future
changes from the observed fields.
\begin{table}[t]
\centering
\caption{Complete task-level failure analysis. Entries use the same primary/multi-label convention as Table~\ref{tab:app_error_groups}. Scores include the rubric convention for non-applicable dimensions.}
\label{tab:app_error_tasks}
\small
\setlength{\tabcolsep}{3pt}
\begin{adjustbox}{max width=\linewidth}
\begin{tabular*}{\linewidth}{@{\extracolsep{\fill}}lrccccccr@{}}
\toprule
Task & $n$ & E1 & E2 & E3 & E4 & E5 & E6 & Mean score \\
\midrule
C01 & 5 & 3/4 & 2/4 & 0/1 & 0/0 & 0/0 & 0/0 & 7.80 \\
C02 & 10 & 3/7 & 7/10 & 0/1 & 0/0 & 0/2 & 0/0 & 7.10 \\
C03 & 6 & 3/4 & 2/2 & 0/2 & 0/1 & 1/1 & 0/0 & 6.00 \\
C04 & 8 & 6/7 & 1/1 & 0/0 & 1/3 & 0/0 & 0/0 & 7.12 \\
C05 & 5 & 2/3 & 3/5 & 0/1 & 0/0 & 0/0 & 0/0 & 7.40 \\
C06 & 10 & 4/5 & 6/10 & 0/2 & 0/0 & 0/3 & 0/0 & 6.90 \\
C07 & 5 & 2/3 & 3/4 & 0/4 & 0/0 & 0/0 & 0/0 & 5.60 \\
C08 & 8 & 2/4 & 3/5 & 0/2 & 0/0 & 3/7 & 0/0 & 4.62 \\
S01 & 8 & 0/0 & 8/8 & 0/0 & 0/0 & 0/8 & 0/0 & 7.38 \\
S02 & 10 & 2/4 & 6/9 & 0/1 & 0/0 & 2/6 & 0/0 & 5.90 \\
S03 & 5 & 0/1 & 5/5 & 0/2 & 0/0 & 0/0 & 0/0 & 6.00 \\
S04 & 5 & 3/4 & 0/0 & 2/5 & 0/1 & 0/0 & 0/0 & 5.80 \\
S05 & 8 & 2/3 & 6/8 & 0/5 & 0/0 & 0/0 & 0/0 & 4.88 \\
S06 & 8 & 0/2 & 4/8 & 0/0 & 0/0 & 4/8 & 0/0 & 6.00 \\
S07 & 5 & 2/5 & 3/4 & 0/3 & 0/0 & 0/0 & 0/0 & 4.80 \\
S08 & 5 & 2/4 & 0/3 & 3/3 & 0/1 & 0/0 & 0/0 & 5.20 \\
O01 & 8 & 6/6 & 2/3 & 0/8 & 0/0 & 0/0 & 0/0 & 5.50 \\
O02 & 5 & 5/5 & 0/0 & 0/2 & 0/2 & 0/0 & 0/0 & 6.00 \\
O03 & 8 & 4/5 & 4/4 & 0/7 & 0/0 & 0/0 & 0/0 & 5.12 \\
O04 & 10 & 1/1 & 9/10 & 0/2 & 0/0 & 0/7 & 0/0 & 6.10 \\
O05 & 5 & 3/3 & 2/4 & 0/1 & 0/0 & 0/0 & 0/0 & 4.80 \\
O06 & 5 & 0/3 & 5/5 & 0/3 & 0/0 & 0/0 & 0/1 & 4.80 \\
O07 & 5 & 2/3 & 2/5 & 0/2 & 0/0 & 0/0 & 1/1 & 4.00 \\
O08 & 5 & 2/2 & 3/4 & 0/4 & 0/0 & 0/0 & 0/0 & 4.60 \\
\bottomrule
\end{tabular*}
\end{adjustbox}
\end{table}
\section{Additional Model ANALYSES}
\subsection{ViT Baseline}
\label{app:vit_baseline}
\paragraph{Purpose of the ViT baseline.}
We use the supervised ViT baseline as a sanity check to verify that the rendered physical fields contain learnable visual-physical signals.
Each task has an independently supervised visual classifier, including the six
tasks excluded from MLLM post-training. This baseline tests predictive signal
under task-specific learning. High accuracy can establish learnable associations
without demonstrating a correct physical reasoning process.
\paragraph{Training details.}
We train a separate ViT classifier for each of the 24 tasks, using 250
training examples per task. For the 18 training-covered tasks, these
are the same examples used for answer-only SFT. The ViT classifiers
are also trained on the six held-out tasks.
The ViT encoder divides each image into patches of $16\times16$ pixels, and uses embedding dimension 256,
six Transformer layers, eight attention heads, an MLP expansion
ratio of 4, and dropout 0.1. Frame or view embeddings are averaged
within each case. For paired inputs, let $h_1$ and $h_2$ denote the
representations of the two cases. An MLP classifier predicts the
answer from their concatenated features
$[h_1;h_2;h_1-h_2;h_1\odot h_2]$, where $\odot$ denotes
element-wise multiplication.
The default configuration uses $384\times384$ RGB inputs and
AdamW with learning rate $3\times10^{-4}$, weight decay $10^{-4}$,
batch size 4, and 50 training epochs. The learning rate follows
a cosine schedule to $10^{-6}$.
Since each task is formulated
as a classification problem over answer options, training
minimizes cross-entropy loss.
\paragraph{Per-task performance.}
Table~\ref{tab:app_vit} reports chance-normalized scores for
all 24 tasks. The results show learnable signals across controlled, simulated,
and observed fields, with substantial variation between tasks. For example,
the classifier reaches perfect accuracy on one-dimensional diffusion comparison,
radiative-layer cooling-time comparison, NASA-CRM, AirCraft drag-coefficient
comparison, and stationary-cylinder Reynolds comparison. Performance is lower
on two-dimensional diffusion and directional-advection comparisons, showing
that learnability varies with the target and visual representation.
\begin{table}[t]
\centering
\caption{Per-task chance-normalized ViT accuracy on PhysFieldBench. Total is the mean across all 24 tasks.}
\label{tab:app_vit}
\small
\begin{tabular*}{\linewidth}{@{\extracolsep{\fill}}lrlr@{}}
\toprule
Task & Norm. & Task & Norm. \\
\midrule
C01 & 85.00  & S05 & 28.00  \\
C02 & 70.00  & S06 & 100.00 \\
C03 & 100.00 & S07 & 100.00 \\
C04 & 45.71  & S08 & 92.00  \\
C05 & 73.33  & O01 & 100.00 \\
C06 & 26.00  & O02 & 80.00  \\
C07 & 25.00  & O03 & 95.00  \\
C08 & 13.33  & O04 & 97.77  \\
S01 & 80.00  & O05 & 64.00  \\
S02 & 42.23  & O06 & 84.00  \\
S03 & 100.00 & O07 & 72.00  \\
S04 & 68.00  & O08 & 48.00  \\
\midrule
\multicolumn{3}{@{}l}{\textbf{Total}} & \textbf{70.39} \\
\bottomrule
\end{tabular*}
\end{table}
\subsection{SFT Data Scaling}
\paragraph{Scaling setup.}
We compare answer-only SFT with 450, 900, 2,250, and 4,500 training
examples, corresponding to 10\%, 20\%, 50\%, and 100\% of the maximum
training size. Each setting uses equal quotas across the eighteen
training-covered tasks and balances answer options.
All models follow the same training recipe.
\paragraph{Scaling results.}
Table~\ref{tab:app_scaling} shows that overall normalized accuracy
increases from 1.68 to 47.65 as supervision grows. The largest gain
occurs between 900 and 2,250 examples, from 4.06 to 45.90, while doubling
the data to 4,500 yields a further gain of only 1.75 points. These
results suggest that SFT gains in physical-field understanding depend
strongly on the amount of supervision, with diminishing gains at larger
scales. Limited training data may be insufficient for learning reliable
associations between physical-field structures and the target mechanisms,
parameters, or outcomes.
Increasing the training set from 2,250 to 4,500 examples improves
covered-task performance from 52.59 to 58.15, while held-out performance
declines from 25.83 to 16.17. Thus, stronger performance on covered tasks
does not necessarily translate into better transfer.
\begin{table}[t]
\centering
\caption{Answer-only SFT data scaling. Fractions are relative to the maximum training size of 4,500 examples. Scores are chance-normalized accuracy averaged over the eighteen training-covered tasks, six held-out tasks, or all 24 tasks. Each row represents one training run.}
\label{tab:app_scaling}
\small
\begin{tabular*}{\linewidth}{@{\extracolsep{\fill}}lrrrr@{}}
\toprule
Fraction & Training examples & Covered & Held-out & All \\
\midrule
10\%  & 450   & 2.18  & 0.17  & 1.68  \\
20\%  & 900   & 2.96  & 7.33  & 4.06  \\
50\%  & 2,250 & 52.59 & 25.83 & 45.90 \\
100\% & 4,500 & 58.15 & 16.17 & 47.65 \\
\bottomrule
\end{tabular*}
\end{table}
\subsection{Prompting and Reasoning Ablations}
\paragraph{Few-shot prompting.}
As Table~\ref{tab:app_fewshot} shows, two-shot prompting provides no consistent benefit for Gemini 3 Flash
on the eighteen training-covered tasks.
Accuracy improves on six tasks, remains unchanged on four, and declines
on eight, reducing the mean normalized score from 24.67 to 23.03.
The largest gain occurs in cylinder Reynolds-number comparison (O01),
while 1D diffusion comparison (C03) shows the largest decline.
These results suggest that the benefit of demonstrations is task-dependent
and that two examples alone do not reliably improve physical-field reasoning
under the evaluated setting.
\begin{table}[t]
\centering
\caption{Zero-shot and two-shot Gemini 3 Flash chance-normalized scores on training-covered tasks. $\Delta$ is Two-shot minus Zero-shot in normalized-score points. Mean is the average across tasks.}
\label{tab:app_fewshot}
\small
\begin{tabular*}{\linewidth}{@{\extracolsep{\fill}}lrrr@{}}
\toprule
Task & Zero-shot & Two-shot & $\Delta$ \\
\midrule
C01 & 42.50 & 47.50 & +5.00 \\
C03 & 70.00 & 50.00 & -20.00 \\
C04 & 17.14 & 11.43 & -5.71 \\
C05 & 56.67 & 46.67 & -10.00 \\
C06 & 24.00 & 28.00 & +4.00 \\
C08 & 16.67 & 13.33 & -3.33 \\
S01 & 28.89 & 35.56 & +6.67 \\
S02 & 22.23 & 24.44 & +2.21 \\
S03 & 40.00 & 25.00 & -15.00 \\
S04 & 32.00 & 24.00 & -8.00 \\
S05 & -4.00 & 0.00 & +4.00 \\
S06 & 9.33 & 6.67 & -2.67 \\
O01 & 5.00 & 25.00 & +20.00 \\
O02 & 5.00 & 5.00 & +0.00 \\
O04 & -13.33 & -20.00 & -6.67 \\
O05 & 24.00 & 24.00 & +0.00 \\
O06 & 36.00 & 36.00 & +0.00 \\
O08 & 32.00 & 32.00 & +0.00 \\
\midrule
Mean & 24.67 & 23.03 & -1.64 \\
\bottomrule
\end{tabular*}
\end{table}
\section{Full Evaluation Results}
\subsection{Per-task Results}
We report per-task chance-normalized scores of all models and post-training variants evaluated in the main text. ViT is supervised separately for each task. The post-training
block contains Qwen3-VL-8B-Thinking and its answer-only SFT, CoT-SFT, and RL
variants. Task identifiers follow the benchmark inventory, and $\dagger$ marks
the six tasks excluded from MLLM post-training.
\paragraph{Controlled equation fields.}
Table~\ref{tab:app_full_controlled} reports results for C01--C08 and shows substantial differences between
recognizing mechanisms and estimating their coefficients. GPT-5.5 has the
highest zero-shot domain mean (43.5), but the best zero-shot score on joint
2D advection-coefficient comparison (C08) is only 20.00. Answer-only SFT
reaches 100.00 on covered 1D diffusion comparison (C03), yet scores $-5.00$
on its held-out 2D counterpart (C07). This contrast illustrates the limited
transfer of learned parameter comparisons across spatial dimensions.
\begin{table}[tbp]
\centering
\small
\caption{Chance-normalized scores on controlled equation fields. Mean averages C01--C08. $\dagger$ marks tasks excluded from MLLM post-training. ViT uses per-task supervision. Within the zero-shot MLLMs, column-best scores are shown in \textbf{bold}. Within the Qwen3-VL-8B-Thinking variants, column-best scores are \underline{underlined}. Ties are all marked.}
\label{tab:app_full_controlled}
\setlength{\tabcolsep}{2pt}
\begin{tabular*}{\linewidth}{@{\extracolsep{\fill}}lrrrrrrrrr@{}}
\toprule
Model & C01 & C02$^{\dagger}$ & C03 & C04 & C05 & C06 & C07$^{\dagger}$ & C08 & Mean \\
\midrule
\multicolumn{10}{@{}l}{\textit{Supervised visual baseline}} \\
ViT & 85.0 & 70.0 & 100.0 & 45.7 & 73.3 & 26.0 & 25.0 & 13.3 & 54.8 \\
\midrule
\multicolumn{10}{@{}l}{\textit{Zero-shot MLLMs}} \\
Qwen3-VL-8B-Instruct & 0.0 & 0.0 & -20.0 & -2.9 & 10.0 & 14.0 & -15.0 & 3.3 & -1.3 \\
LLaVA-OneVision-2-8B & 7.5 & 0.0 & -5.0 & -2.9 & 3.3 & -2.0 & 0.0 & 0.0 & 0.1 \\
Intern-S1-Pro & 25.0 & 28.0 & 10.0 & -5.7 & 3.3 & 4.0 & \textbf{50.0} & 0.0 & 14.3 \\
Kimi K2.6 & 50.0 & 36.0 & 60.0 & 5.7 & 43.3 & 24.0 & 25.0 & 3.3 & 30.9 \\
GLM-5V-Turbo & 27.5 & 32.0 & 65.0 & -2.9 & 33.3 & 8.0 & 15.0 & 6.7 & 23.1 \\
Qwen3.6-Plus & 52.5 & 28.0 & 80.0 & 5.7 & 46.7 & 14.0 & 20.0 & \textbf{20.0} & 33.4 \\
Doubao-Seed-2.0-Pro & 37.5 & 28.0 & 60.0 & -8.6 & 46.7 & 16.0 & 10.0 & 3.3 & 24.1 \\
Claude Opus 4.7 & 60.0 & 32.0 & 0.0 & 0.0 & \textbf{60.0} & \textbf{34.0} & -5.0 & 13.3 & 24.3 \\
Gemini 3 Flash & 42.5 & 26.0 & 70.0 & 17.1 & 56.7 & 24.0 & -5.0 & 16.7 & 31.0 \\
Gemini 3.1 Pro & \textbf{65.0} & 44.0 & \textbf{85.0} & 5.7 & \textbf{60.0} & 18.0 & 20.0 & 6.7 & 38.1 \\
Gemini 3.5 Flash & 60.0 & \textbf{46.0} & 80.0 & 28.6 & \textbf{60.0} & 20.0 & 25.0 & 13.3 & 41.6 \\
GPT-5.5 & \textbf{65.0} & \textbf{46.0} & 80.0 & \textbf{45.7} & 53.3 & 28.0 & 20.0 & 10.0 & \textbf{43.5} \\
\midrule
\multicolumn{10}{@{}l}{\textit{Qwen3-VL-8B-Thinking variants}} \\
Base & 5.0 & -14.0 & -70.0 & -2.9 & 13.3 & 4.0 & 10.0 & 0.0 & -6.8 \\
Answer-only SFT & \underline{90.0} & \underline{44.0} & \underline{100.0} & \underline{14.3} & \underline{63.3} & \underline{44.0} & -5.0 & \underline{26.7} & \underline{47.2} \\
CoT SFT & 35.0 & 28.0 & 95.0 & 0.0 & 36.7 & 10.0 & 15.0 & 0.0 & 27.5 \\
CoT SFT + RL & 60.0 & 34.0 & 80.0 & 0.0 & 30.0 & 12.0 & \underline{35.0} & 3.3 & 31.8 \\
\bottomrule
\end{tabular*}
\end{table}
\paragraph{Simulated physical fields.}
As shown in Table~\ref{tab:app_full_simulated}, model strengths vary across the physical systems represented by S01--S08. Doubao achieves the highest zero-shot
domain mean (30.9), while the strongest zero-shot scores for car lift
comparison (S05) and joint Mach--angle inference (S06) are only 20.0 and
30.7. Answer-only SFT reaches 100.0 on cooling-time comparison (S03) and
97.3 on NASA-CRM (S06), but only 16.0 and 8.0 on the held-out aircraft coefficient tasks (S07--S08). Supervision therefore improves several covered
systems substantially without ensuring transfer to new aerodynamic tasks.
\begin{table}[tbp]
\centering
\small
\caption{Chance-normalized scores on simulated physical fields. Mean averages S01--S08. $\dagger$ marks tasks excluded from MLLM post-training. ViT uses per-task supervision. Within the zero-shot MLLMs, column-best scores are shown in \textbf{bold}. Within the Qwen3-VL-8B-Thinking variants, column-best scores are \underline{underlined}. Ties are all marked.}
\label{tab:app_full_simulated}
\setlength{\tabcolsep}{2pt}
\begin{tabular*}{\linewidth}{@{\extracolsep{\fill}}lrrrrrrrrr@{}}
\toprule
Model & S01 & S02 & S03 & S04 & S05 & S06 & S07$^{\dagger}$ & S08$^{\dagger}$ & Mean \\
\midrule
\multicolumn{10}{@{}l}{\textit{Supervised visual baseline}} \\
ViT & 80.0 & 42.2 & 100.0 & 68.0 & 28.0 & 100.0 & 100.0 & 92.0 & 76.3 \\
\midrule
\multicolumn{10}{@{}l}{\textit{Zero-shot MLLMs}} \\
Qwen3-VL-8B-Instruct & 0.0 & 0.0 & -30.0 & -20.0 & -8.0 & 4.0 & 4.0 & 16.0 & -4.3 \\
LLaVA-OneVision-2-8B & 2.2 & 0.0 & 0.0 & -20.0 & -12.0 & -9.3 & 0.0 & -20.0 & -7.4 \\
Intern-S1-Pro & 17.8 & 15.6 & -30.0 & 4.0 & \textbf{20.0} & -4.0 & 48.0 & 32.0 & 12.9 \\
Kimi K2.6 & 33.3 & 20.0 & -5.0 & \textbf{48.0} & 4.0 & 4.0 & 48.0 & 28.0 & 22.5 \\
GLM-5V-Turbo & 15.6 & 20.0 & -45.0 & \textbf{48.0} & 0.0 & 6.7 & 20.0 & 28.0 & 11.7 \\
Qwen3.6-Plus & 33.3 & 28.9 & -50.0 & 36.0 & 0.0 & \textbf{30.7} & \textbf{64.0} & 40.0 & 22.9 \\
Doubao-Seed-2.0-Pro & 26.7 & \textbf{40.0} & 35.0 & 20.0 & 12.0 & 17.3 & 48.0 & 48.0 & \textbf{30.9} \\
Claude Opus 4.7 & -2.2 & -13.3 & 5.0 & -20.0 & -8.0 & 22.7 & 20.0 & 4.0 & 1.0 \\
Gemini 3 Flash & 28.9 & 22.2 & \textbf{40.0} & 32.0 & -4.0 & 9.3 & 60.0 & 56.0 & 30.6 \\
Gemini 3.1 Pro & 35.6 & 17.8 & -55.0 & 40.0 & \textbf{20.0} & 9.3 & 52.0 & \textbf{60.0} & 22.5 \\
Gemini 3.5 Flash & 40.0 & 26.7 & -40.0 & 24.0 & 16.0 & 17.3 & 52.0 & 52.0 & 23.5 \\
GPT-5.5 & \textbf{46.7} & 13.3 & -30.0 & 20.0 & -8.0 & 22.7 & 60.0 & 48.0 & 21.6 \\
\midrule
\multicolumn{10}{@{}l}{\textit{Qwen3-VL-8B-Thinking variants}} \\
Base & 4.4 & -4.4 & -35.0 & 12.0 & -24.0 & 17.3 & -32.0 & -20.0 & -10.2 \\
Answer-only SFT & \underline{75.6} & \underline{35.6} & \underline{100.0} & 28.0 & 24.0 & \underline{97.3} & 16.0 & 8.0 & \underline{48.1} \\
CoT SFT & 13.3 & 26.7 & 20.0 & \underline{40.0} & \underline{32.0} & 14.7 & 20.0 & 16.0 & 22.8 \\
CoT SFT + RL & 53.3 & 20.0 & 80.0 & 36.0 & 24.0 & 49.3 & \underline{32.0} & \underline{20.0} & 39.3 \\
\bottomrule
\end{tabular*}
\end{table}
\paragraph{Observed physical fields.}
Table~\ref{tab:app_full_observed} reports evaluation results for O01--O08. RL achieves the highest MLLM domain mean (56.0), compared with 47.7 for
answer-only SFT and 31.0 for CoT-SFT. Its gains over CoT-SFT are pronounced
for cylinder Reynolds-number comparison (O01) and joint
foil-parameter inference (O04). Transfer remains uneven.
RL improves held-out controlled-cylinder comparison (O03) from 15.0 to
35.0, but improves held-out cyclone forecasting (O07) only from 40.0
to 44.0. ViT's domain mean of 80.1 further indicates substantial room
for MLLMs to exploit the information in observed physical fields.
\begin{table}[tbp]
\centering
\small
\caption{Chance-normalized scores on observed physical fields. Mean averages O01--O08. $\dagger$ marks tasks excluded from MLLM post-training. ViT uses per-task supervision. Within the zero-shot MLLMs, column-best scores are shown in \textbf{bold}. Within the Qwen3-VL-8B-Thinking variants, column-best scores are \underline{underlined}. Ties are all marked.}
\label{tab:app_full_observed}
\setlength{\tabcolsep}{2pt}
\begin{tabular*}{\linewidth}{@{\extracolsep{\fill}}lrrrrrrrrr@{}}
\toprule
Model & O01 & O02 & O03$^{\dagger}$ & O04 & O05 & O06 & O07$^{\dagger}$ & O08 & Mean \\
\midrule
\multicolumn{10}{@{}l}{\textit{Supervised visual baseline}} \\
ViT & 100.0 & 80.0 & 95.0 & 97.8 & 64.0 & 84.0 & 72.0 & 48.0 & 80.1 \\
\midrule
\multicolumn{10}{@{}l}{\textit{Zero-shot MLLMs}} \\
Qwen3-VL-8B-Instruct & 0.0 & -10.0 & 35.0 & 0.0 & 4.0 & 28.0 & \textbf{52.0} & 0.0 & 13.6 \\
LLaVA-OneVision-2-8B & 0.0 & 0.0 & 0.0 & 0.0 & 8.0 & 4.0 & 0.0 & 0.0 & 1.5 \\
Intern-S1-Pro & -5.0 & -5.0 & 15.0 & 6.7 & 8.0 & 24.0 & 24.0 & \textbf{44.0} & 14.0 \\
Kimi K2.6 & 25.0 & -10.0 & 30.0 & -20.0 & \textbf{36.0} & 12.0 & 40.0 & 32.0 & 18.1 \\
GLM-5V-Turbo & 5.0 & 5.0 & 20.0 & \textbf{40.0} & 8.0 & 20.0 & 48.0 & 36.0 & \textbf{22.8} \\
Qwen3.6-Plus & 35.0 & \textbf{20.0} & 20.0 & -6.7 & 16.0 & 20.0 & 28.0 & 40.0 & 21.5 \\
Doubao-Seed-2.0-Pro & 40.0 & 15.0 & 5.0 & -4.4 & 8.0 & 28.0 & 4.0 & 24.0 & 15.0 \\
Claude Opus 4.7 & 5.0 & -5.0 & 5.0 & -8.9 & -8.0 & -4.0 & -4.0 & -16.0 & -4.5 \\
Gemini 3 Flash & 5.0 & 5.0 & 5.0 & -13.3 & 24.0 & \textbf{36.0} & 36.0 & 32.0 & 16.2 \\
Gemini 3.1 Pro & 5.0 & -10.0 & 5.0 & -15.6 & 32.0 & 28.0 & \textbf{52.0} & 36.0 & 16.6 \\
Gemini 3.5 Flash & -30.0 & 0.0 & -35.0 & -15.6 & \textbf{36.0} & \textbf{36.0} & 44.0 & 36.0 & 8.9 \\
GPT-5.5 & \textbf{60.0} & -30.0 & \textbf{45.0} & -13.3 & 20.0 & \textbf{36.0} & 32.0 & 32.0 & 22.7 \\
\midrule
\multicolumn{10}{@{}l}{\textit{Qwen3-VL-8B-Thinking variants}} \\
Base & -10.0 & -35.0 & 0.0 & 0.0 & -24.0 & 16.0 & 28.0 & -12.0 & -4.6 \\
Answer-only SFT & \underline{95.0} & \underline{60.0} & -10.0 & 68.9 & \underline{56.0} & 36.0 & \underline{44.0} & 32.0 & 47.7 \\
CoT SFT & 55.0 & 40.0 & 15.0 & 22.2 & 16.0 & 24.0 & 40.0 & 36.0 & 31.0 \\
CoT SFT + RL & \underline{95.0} & 45.0 & \underline{35.0} & \underline{88.9} & \underline{56.0} & \underline{44.0} & \underline{44.0} & \underline{40.0} & \underline{56.0} \\
\bottomrule
\end{tabular*}
\end{table}
\section{Additional Showcases}
\subsection{Representative Examples}
We present representative examples from the three field
domains to illustrate the visual inputs and inference targets in
PhysFieldBench. Each example includes the field images and ground truth.
\paragraph{Controlled equation fields.}
Figure~\ref{fig:app_representative_examples}(a) illustrates
single-term recognition (C01) through spacetime fields dominated
by reaction, diffusion, linear advection, and nonlinear advection.
The task asks the model to identify the dominant term from five
options: reaction, quadratic reaction, linear advection,
nonlinear advection, and diffusion. These examples illustrate
the distinct field evolution associated with different mechanisms.
\paragraph{Simulated physical fields.}
Figure~\ref{fig:app_representative_examples}(b) presents two
NASA-CRM cases for joint Mach-number and angle-of-attack
comparison (S06). Each case combines multiple views of surface
pressure coefficient and skin-friction magnitude. The task
requires comparing both flight parameters from these complementary
fields. The correct answer is that Case 1 has a lower Mach number
and a higher angle of attack.
\paragraph{Observed physical fields.}
Figure~\ref{fig:app_representative_examples}(c) shows paired
SEVIR scenes with VIS, IR069, and IR107 channels (O08).
The task asks which scene has the larger current strong-VIL
coverage. The radar-derived target areas are 4,458 and 1,874 pixels,
giving answer that Case 1 has the larger coverage. Only the satellite channels are provided
as visual inputs.
\begin{figure}[htbp]
\centering
\includegraphics[width=\linewidth]{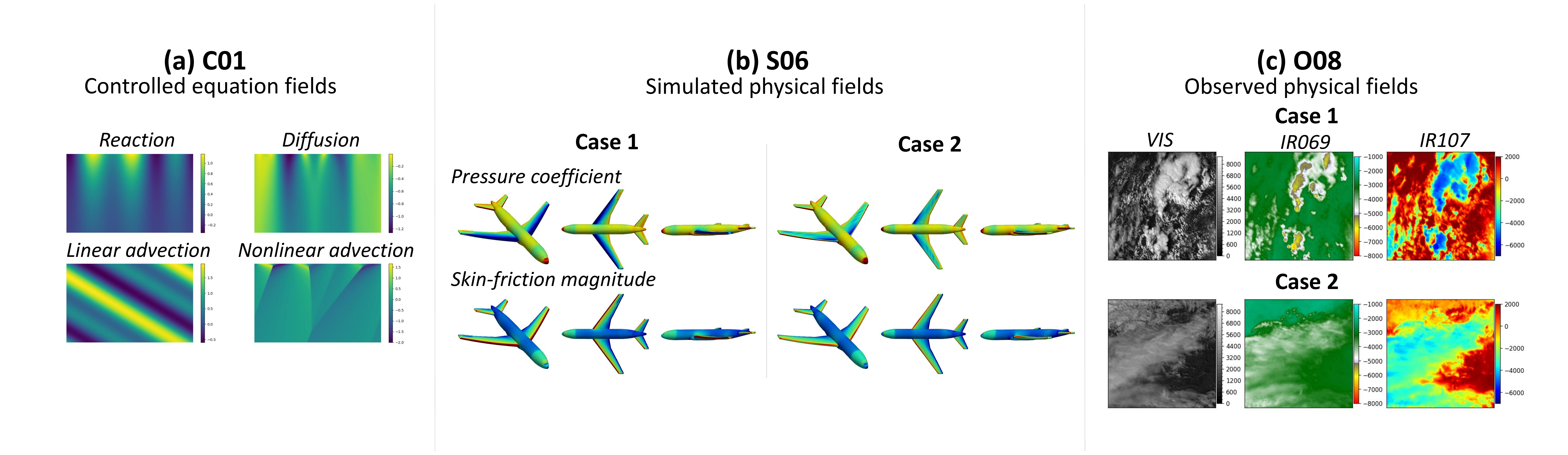}
\caption{Representative examples from the three physical-field
domains. (a) C01: spacetime fields illustrating four dominant
mechanisms, with time increasing downward. (b) S06: paired
NASA-CRM pressure and skin-friction fields for joint Mach-number
and angle-of-attack comparison.
(c) O08: paired satellite observations for radar-derived
strong-VIL coverage comparison.
Mechanism labels in (a) identify the examples for illustration
and are not supplied to the evaluated model.}
\label{fig:app_representative_examples}
\end{figure}
\subsection{Success Cases}
\paragraph{Correct physical mapping.}
The cooling-time example in Figure~\ref{fig:app_showcase_success} (left) illustrates a successful link between density evolution and radiative
cooling. Case 2 develops dense structures reaching approximately 170,
well above its initial high-density phase of about 90, whereas Case 1
remains within its initial density range. The model relates this contrast
to stronger cooling-induced condensation in Case 2 and correctly selects
Case 1 as having the longer cooling time. Comparing each trajectory
against its own initial density range helps distinguish physical change
from differences in color scaling.
\paragraph{Multi-cue integration.}
In the cyclone-size example Figure~\ref{fig:app_showcase_success} (right), the
model combines the broad outer cloud bands visible in infrared imagery
with the extended curved precipitation structure in the passive-microwave
channel. Both cues support a larger spatial extent for Case 1, leading
to the correct answer. The explanation focuses on the extent of the
storm structure across channels, rather than relying solely on the
appearance of its convective core.
\begin{figure}[htbp]
\centering
\includegraphics[width=\linewidth]{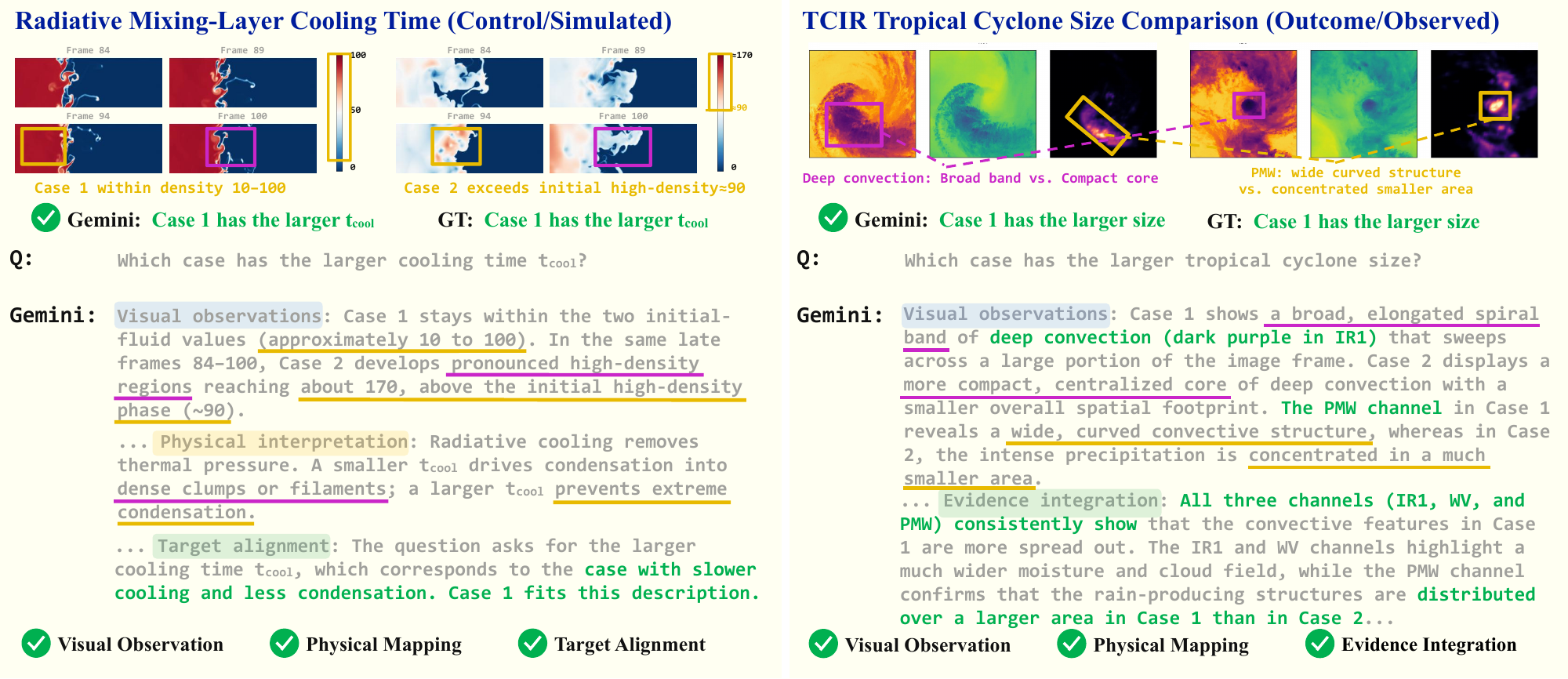}
\caption{Successful physical-field reasoning by Gemini 3 Flash. In cooling-time
comparison (S03), density growth relative to the initial state supports
the correct cooling-time ordering. In cyclone-size comparison (O05),
cloud-band and precipitation extent jointly support the correct size
ordering.}
\label{fig:app_showcase_success}
\end{figure}
\subsection{Failure Cases}
\begin{figure}[htbp]
\centering
\includegraphics[width=\linewidth]{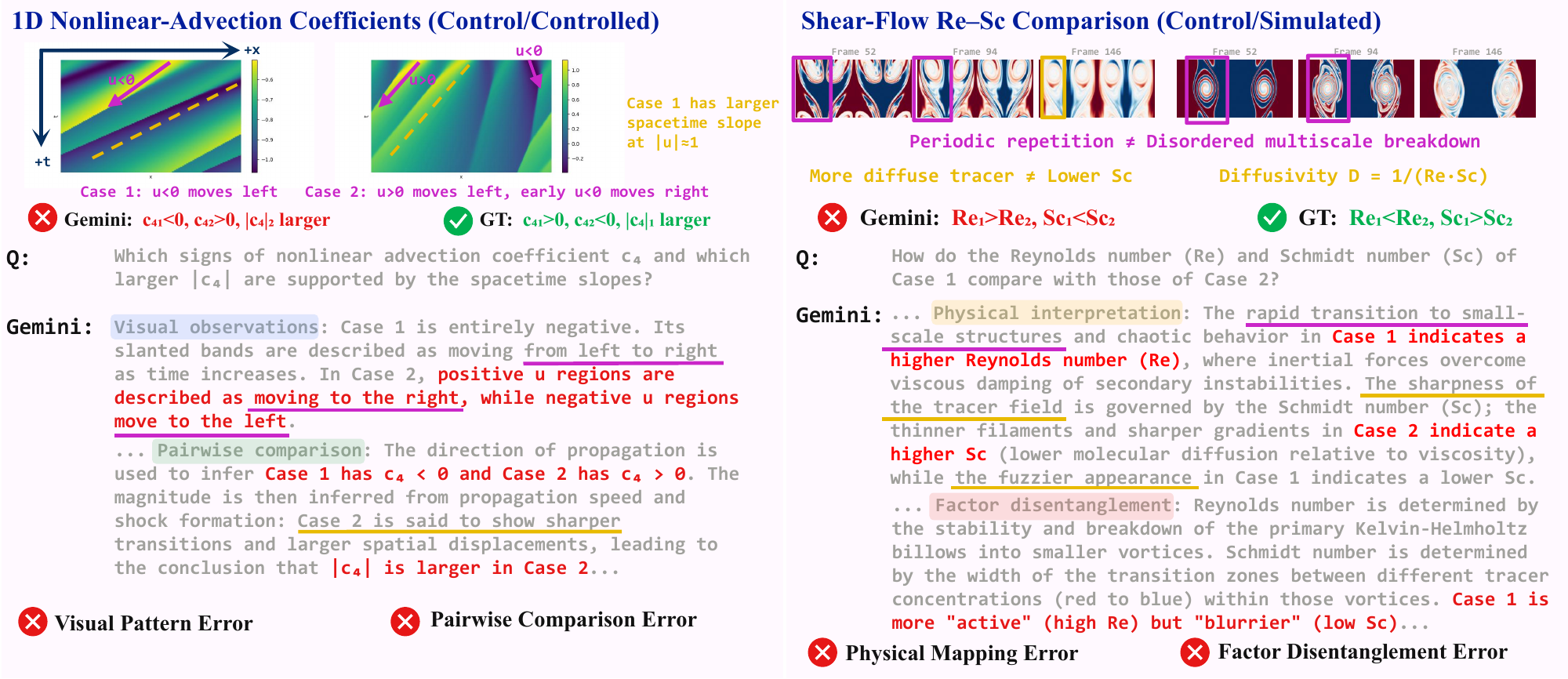}
\caption{Failed physical-field reasoning by Gemini.
In nonlinear-advection comparison (C04), misread propagation
directions lead to reversing the coefficient signs
and magnitude ordering.
In shear-flow comparison (S02), the model mistakes periodic
structures for chaotic breakdown and associates tracer blurring
directly with lower Sc.}
\label{fig:app_showcase_failure}
\end{figure}
\paragraph{Visual pattern failure.}
In Figure~\ref{fig:app_showcase_failure} (left), the model reverses
the propagation directions inferred from the spacetime heatmaps,
where time increases downward. Since the nonlinear-advection
velocity $c_4u$ depends on both the coefficient and field signs,
these misread directions lead to incorrect coefficient signs.
The model also reverses the magnitude ordering.
\paragraph{Physical mapping failure.}
In Figure~\ref{fig:app_showcase_failure} (right), the model interprets
the repeated coherent vortices in Case 1 as disordered multiscale
breakdown and infers a larger Reynolds number. The visible periodic
organization does not support this interpretation, illustrating
an incorrect mapping from spatial complexity to flow instability.
\paragraph{Factor disentanglement failure.}
The shear-flow response further attributes the more diffuse tracer
in Case 1 to a lower Schmidt number. However, tracer diffusivity
depends jointly on both parameters through $D=1/(Re \cdot Sc)$.
Here, Case 1 has higher Sc but also higher diffusivity because
its Re is substantially lower. Thus, tracer blurring alone cannot
determine the Sc ordering when Re also varies. The model reverses
both parameter orderings.
\section{Limitations}
\textbf{From static field inference to interactive scientific agents.}
PhysFieldBench evaluates whether models can infer mechanisms, control variables, and outcomes from preselected field observations. Although some inputs depict temporal evolution, each task requires a single response: models cannot request additional evidence, test a hypothesis by changing simulation conditions, or revise an inference in response to feedback. The benchmark therefore does not measure performance in closed-loop scientific workflows. A future extension could allow agents to select measurements or views, intervene on controllable conditions where feasible, and use the resulting observations to update their hypotheses and guide subsequent decisions.
\newpage
\section{Task Prompts}
\label{app:exact_prompts}
\subsection{\texorpdfstring{C01: PDEFormer1D: Single Dominant Mechanism Identification}{C01}}
\label{app:prompt_C01}
\begin{figure}[h]
\centering
\includegraphics[width=\linewidth]{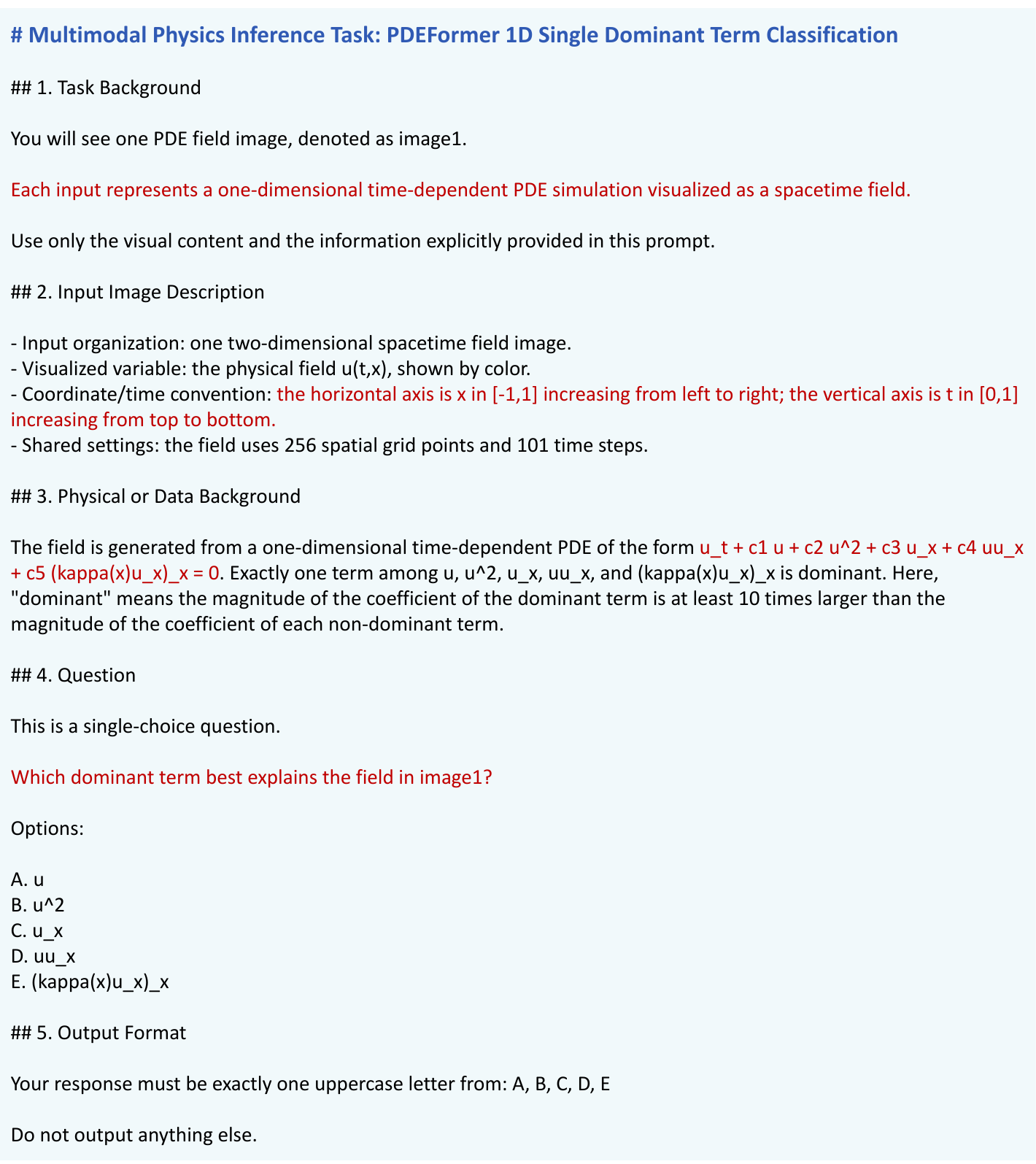}
\caption{Exact evaluation prompt for C01. Given a single 1D spacetime
field, the model identifies the dominant equation term from five
candidate mechanisms. The coefficient-based definition in the prompt specifies the candidate-selection criterion; all retained cases additionally undergo the mechanism-specific manual verification described in Table~\ref{tab:appendix_pair_rules}.}
\label{fig:C01_prompt}
\end{figure}
\clearpage
\subsection{\texorpdfstring{S06: NASA-CRM: Mach Number \& Angle-of-Attack Comparison}{S06}}
\label{app:prompt_S06}
\begin{figure}[h]
\centering
\includegraphics[width=\linewidth]{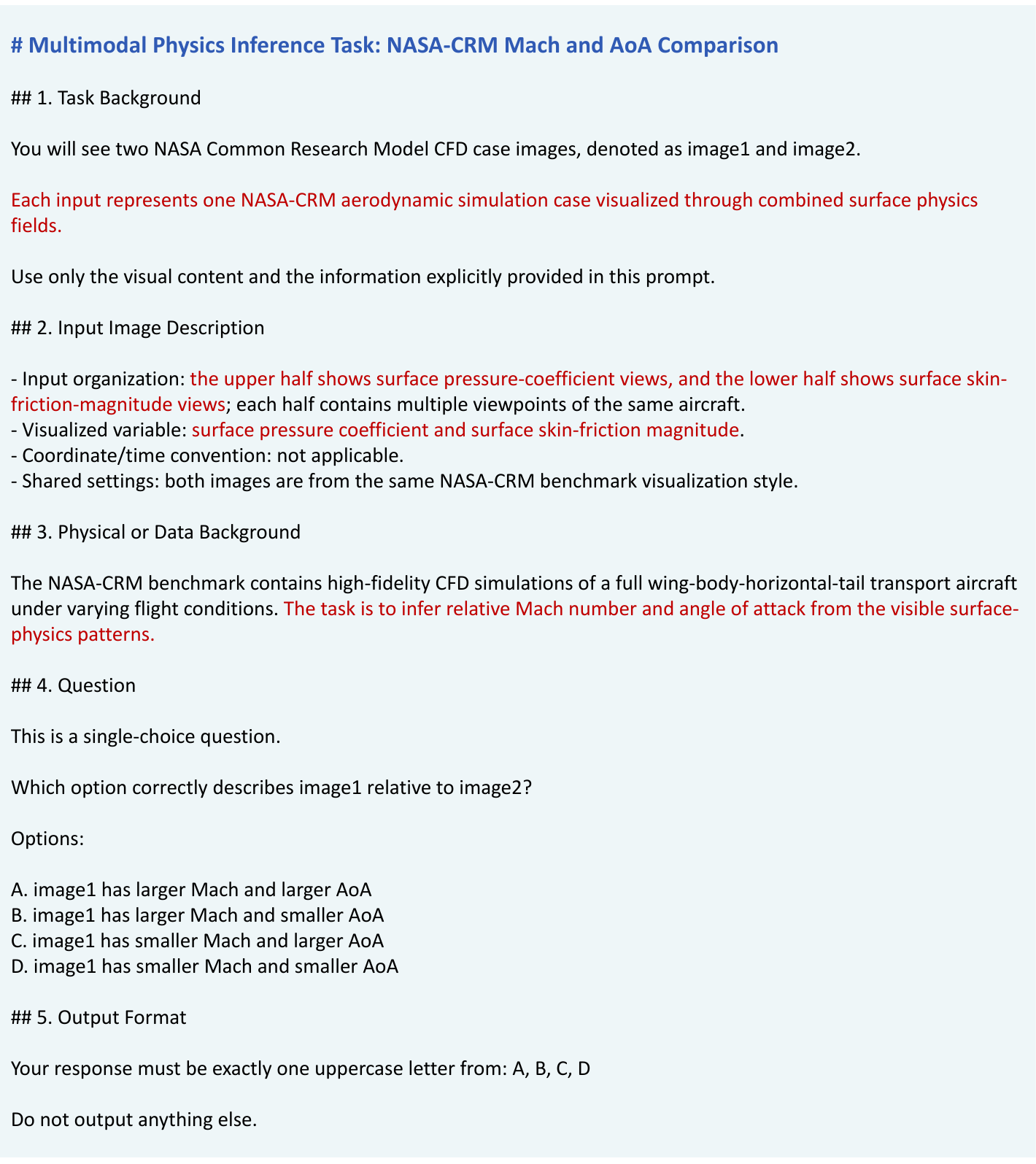}
\caption{Exact evaluation prompt for S06. Given two NASA-CRM cases
visualized through multi-view pressure-coefficient and skin-friction
fields, the model jointly compares their Mach numbers and angles of
attack.}
\label{fig:S06_prompt}
\end{figure}
\clearpage
\subsection{\texorpdfstring{O08: SEVIR: High-VIL Coverage Comparison}{O08}}
\label{app:prompt_O08}
\begin{figure}[h]
\centering
\includegraphics[width=\linewidth]{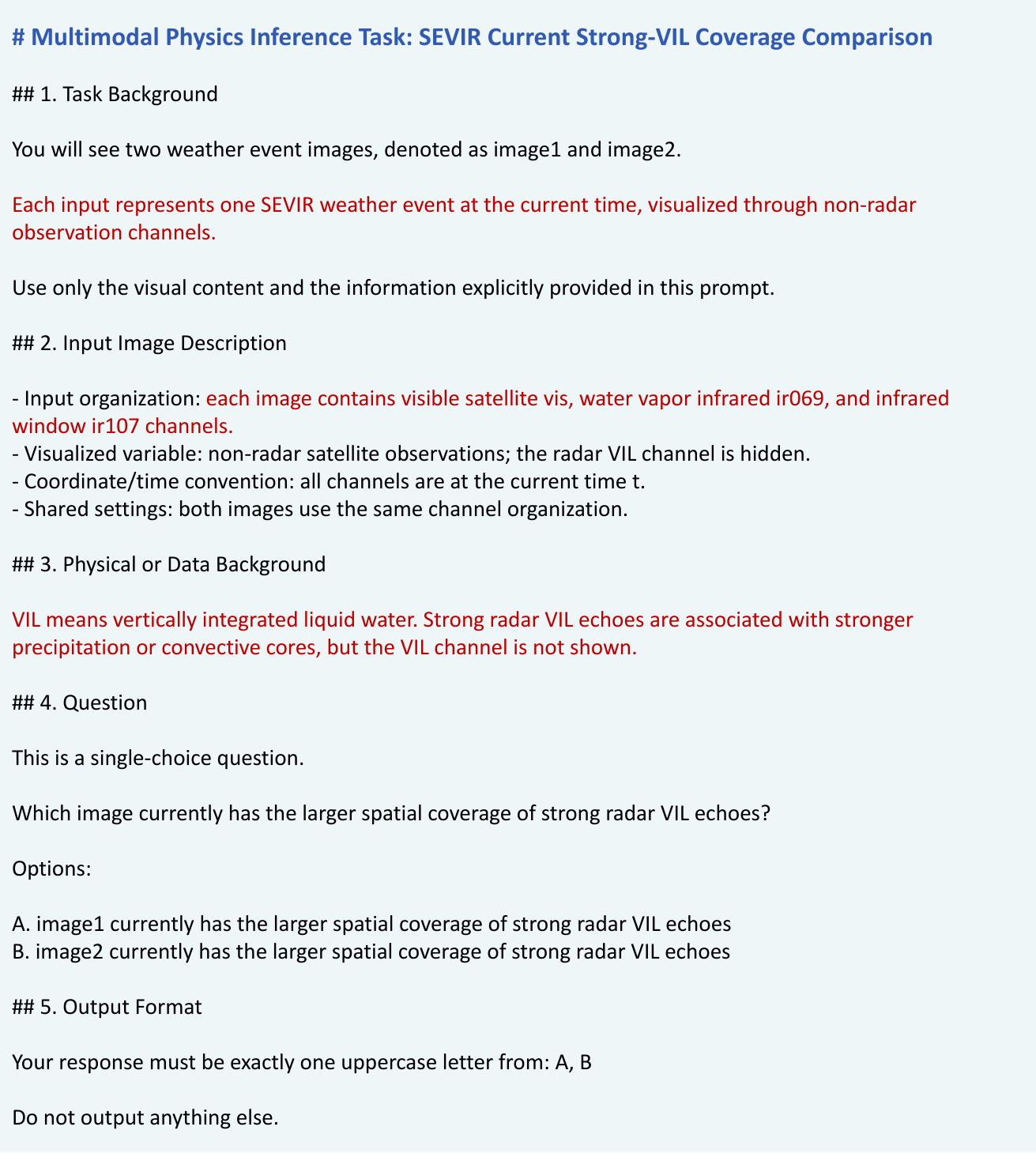}
\caption{Exact evaluation prompt for O08. Given paired VIS, IR069,
and IR107 satellite observations, the model identifies the scene
with larger current strong-VIL coverage. The radar-derived VIL field
is used only to construct the label and is not provided as input.}
\label{fig:O08_prompt}
\end{figure}
\clearpage
\end{document}